%% file: arXiv_main.tex
\documentclass[10pt,twocolumn,letterpaper]{article}

\usepackage[pagenumbers]{wacv} 
\usepackage[accsupp]{axessibility} 

\usepackage{graphicx}
\usepackage{amsmath}
\usepackage{amssymb}
\usepackage{booktabs}

\usepackage{multirow}
\usepackage{tabularx}

\usepackage{changepage}
\usepackage{listings}
\usepackage{pifont}
\newcommand{\xmark}{\ding{55}}%
\usepackage{wrapfig}
\usepackage{floatrow}
\newfloatcommand{capbtabbox}{table}[][\FBwidth]
\usepackage{refcount}
\usepackage{balance}

\usepackage{xr}
\makeatletter
\newcommand*{\addFileDependency}[1]{
  \typeout{(#1)}
  \@addtofilelist{#1}
  \IfFileExists{#1}{}{\typeout{No file #1.}}
}
\makeatother
\newcommand*{\myexternaldocument}[1]{
    \externaldocument{#1}
    \addFileDependency{#1.tex}
    \addFileDependency{#1.aux}
}
\myexternaldocument{supp}

\usepackage[pagebackref,breaklinks,colorlinks]{hyperref}

\usepackage[capitalize]{cleveref}
\crefname{section}{Sec.}{Secs.}
\Crefname{section}{Section}{Sections}
\Crefname{table}{Table}{Tables}
\crefname{table}{Tab.}{Tabs.}

\def\wacvPaperID{401} 
\def\confName{WACV}
\def\confYear{2025}

\begin{document}

\title{Exploiting VLM Localizability and Semantics for \\ Open Vocabulary Action Detection}

\author{
Wentao Bao $^{1*}$, 
Kai Li$^{2}$, 
Yuxiao Chen$^{3*}$, 
Deep Patel$^{2}$, 
Martin Renqiang Min$^{2}$, 
Yu Kong$^{1}$\\
$^{1}$Department of Computer Science and Engineering, Michigan State University \\ 
$^{2}$Machine Learning Department, NEC Laboratories America\\
$^{3}$Department of Computer Science, Rutgers University\\
{\tt\small \{baowenta,yukong\}@msu.edu, yc984@cs.rutgers.edu,} \\
{\tt\small \{kaili,dpatel,renqiang\}@nec-labs.com}
}
\maketitle

\def\thefootnote{*}\footnotetext{Most of this work was done during the internship of Wentao Bao at NEC Labs America and finished at Michigan State University.}

\begin{abstract}
   Action detection aims to detect (recognize and localize) human actions spatially and temporally in videos. Existing approaches focus on the closed-set setting where an action detector is trained and tested on videos from a fixed set of action categories. However, this constrained setting is not viable in an open world where test videos inevitably come beyond the trained action categories. In this paper, we address the practical yet challenging Open-Vocabulary Action Detection (OVAD) problem. It aims to detect any action in test videos while training a model on a fixed set of action categories. To achieve such an open-vocabulary capability, we propose a novel method OpenMixer that exploits the inherent semantics and localizability of large vision-language models (VLM) within the family of query-based detection transformers (DETR). Specifically, the OpenMixer is developed by spatial and temporal OpenMixer blocks (S-OMB and T-OMB), and a dynamically fused alignment (DFA) module. The three components collectively enjoy the merits of strong generalization from pre-trained VLMs and end-to-end learning from DETR design. Moreover, we established OVAD benchmarks under various settings, and the experimental results show that the OpenMixer performs the best over baselines for detecting seen and unseen actions. We release the codes, models, and dataset splits at {\small{\url{https://github.com/Cogito2012/OpenMixer}}}.
\end{abstract}

\section{Introduction}
\label{sec:intro}

Action Detection (AD) aims to recognize actions and spatially and temporally localize the actors in videos. It plays a vital roles in various applications like video surveillance~\cite{yang2009detecting,yun2019vision,dave_WACV2022}, autonomous driving~\cite{ROAD_TPAMI23}, and sport event analysis~\cite{MultiSports_ICCV21}, and it thus draws increasing attentions in recent literature \cite{ACTDet_ICCV17,ACRN_ECCV18,yowo19,ACAR_CVPR21,WOO_ICCV21,CRCNN_ECCV20,TubeR_CVPR22,STMixer_CVPR23,EVAD_ICCV23}. 

Existing AD methods are mostly developed in a closed-set setting where the models are trained and tested on videos from the same fixed set of action categories. While significant progress has been made over the past few years \cite{WOO_ICCV21,TubeR_CVPR22,STMixer_CVPR23,EVAD_ICCV23}, the assumption that the training and test videos are from the same action categories limits their application to the real world, where test videos could contain actions beyond the pre-defined training categories. For example, a video surveillance system may be able to detect \texttt{fighting}, but other dangerous or suspicious actions like \texttt{shooting} and \texttt{chasing} will not be detected if the system has not been trained with annotated videos from these action categories. In addition, being able to detect actions in an open world facilitates a comprehensive understanding of videos and opens doors to high-level video understanding tasks, like reasoning~\cite{VCR_CVPR19}, forecasting~\cite{Sun_2019_CVPR}, \etc., that usually require detecting various actions in videos.  

This motivates us to investigate Open-Vocabulary Action Detection (\textbf{OVAD}), a task aiming to detect any actions in videos, including both seen categories contained in the training set and unseen categories absent in the training set. However, OVAD is challenging as it requires understanding the human motion dynamics across frames. While motion dynamics modeling has been well studied by the conventional closed-set action detection~\cite{SlowFast_ICCV19,WOO_ICCV21,TubeR_CVPR22,STMixer_CVPR23} that takes advantages of full supervision in training, it is challenging in the open-vocabulary setting since there is no supervision for the unseen action categories.

Recently, harvesting the strong generalization capability of pre-trained large visual-language foundation models (VLMs)~\cite{CLIP_ICML21}, various open-vocabulary approaches have been proposed for image recognition~\cite{Zhang_2021_CVPR}, object detection \cite{bangalath2022bridging,ROViT_CVPR23,CORA_CVPR23,VLDet_ICLR23,FVLM_ICLR23}, and image segmentation~\cite{ZegCLIP_CVPR23,liang2023open}. However, these methods are designed for images and do not consider  temporal dynamics among video frames. In addition, image VLMs such as the CLIP~\cite{CLIP_ICML21} are struggling to capture the action verbs in text and human motion in videos~\cite{Verbs_ICCV23}. This inevitably requires learning the temporal dynamics~\cite{ju2022prompting} or fully fine-tuning~\cite{ViFiCLIP_CVPR23} for recognizing the actions on downstream tasks, which take the risk of poor generalizability to the unseen. 

There are a few seminal works that leverage VLMs for open-vocabulary video understanding, including action recognition~\cite{ni2022expanding,OpenVCLIP_ICML23,STAN_CVPR23,ViFiCLIP_CVPR23} and temporal action localization~\cite{rathod2022open_bmvc22,nag2022zero,UniLoc_ICCV23}. However, for the region-level action detection by VLM, there exists a representation gap between video-level pre-training and the region-level adaptation, which is analogous to the representation gap issue discussed in image-based open-vocabulary object detection literature~\cite{bangalath2022bridging,ROViT_CVPR23,CORA_CVPR23}. Specific to the OVAD task, the representation gap stems from the holistic video-action alignment in pre-training and the downstream region-level sub-tasks, \ie, region-action alignment and action-relevant person localization. The cause of the representation gap can be attributed to their intrinsically different adaptation goals from pre-trained video VLMs, \ie, transferring the \emph{semantics} and \emph{localizability} of VLMs from video to regions for the two sub-tasks, respectively. 

Re-thinking the Transformer-like design of VLMs, we found that the way of using VLM semantic features and the undervalued localizability of VLMs are both critical to the OVAD task. 
First, to transfer the video-level semantics to each region, we propose to learn a set of region-wise queries to decode the temporal dynamics from videos, by using the pre-trained video-level features as adaptive semantics conditions. The updated queries and video-level features are further dynamically fused and aligned with the textual semantics for recognition. Second, to exploit the video VLM localizability for region-wise localization, we learn a set of queries to decode the person boxes starting from the prior locations revealed by the VLM visual attention. 

Specifically, we develop a query-based open-vocabulary action detector, OpenMixer, to detect any video actions in an open vocabulary. It fits in the family of the detection transformers (DETR)~\cite{DETR_ECCV20,SparseDETR_ICLR22,AdaMixer_CVPR22,TubeR_CVPR22,STMixer_CVPR23}. The basic idea is to decouple the action recognition and localization by learning two sets of queries and corresponding decoding modules. Our OpenMixer consists of a spatial OpenMixer Block (S-OMB) for person localization, a temporal OpenMixer Block (T-OMB) for capturing the region-level temporal motion, and a dynamically fused alignment (DFA) for open-vocabulary action recognition. The S-OMB inherits the localizability of VLMs by the text-patch cross-attention, the T-OMB exploits the visual semantic features of VLMs to capture the temporal dynamics, and the DFA dynamically fuses the pre-trained semantics into learnable region-level queries for generalizable recognition. Eventually, our model enjoys the merits of semantics and localizability from VLMs and the end-to-end detection capability from the DETR pipeline. 

In experiments, we set up OVAD benchmarks based on popular action detection datasets and evaluate technically viable baselines, showing the superior performance of OpenMixer. In summary, the contributions are three-fold:
\vspace{-1.5mm}
\begin{itemize}
\itemsep0em
    \item We formulate the task of open-vocabulary action detection (OVAD), which is valuable while challenging even by foundation models.
    \item We develop the OpenMixer model that exploits the semantics and localizability of pre-trained video-language models toward the OVAD task.
    \item We empirically reveal the effectiveness of the proposed modules that show strong generalizability on multiple video action detection datasets.
\end{itemize}

\section{Related Work}
\label{sec:related_work}

\textbf{Spatio-temporal Action Detection.} This task aims to localize human actions spatially and temporally in videos and recognize their actions, which has been a fundamental video understanding topic~\cite{LFB_CVPR19,SlowFast_ICCV19,X3D_CVPR20,Hiera_ICML23}. A line of recent works~\cite{ACRN_ECCV18,yowo19,AIA_ECCV20,ACAR_CVPR21, CycleACR_arXiv23,MRSN_arXiv23} adopts the two-backbone design to separately extract features of the keyframes and the entire video for actor localization and actor-context relation modeling, respectively. Though they are flexible by taking advantage of both image and video backbones for achieving promising performance, the model parameters are redundant in design and heavy to optimize~\cite{EVAD_ICCV23}. With the recent advances in detection transformer (DETR)~\cite{DETR_ECCV20}, end-to-end action detection by a single backbone shows impressive performance and thus becomes a more popular design choice~\cite{VAT_CVPR19,WOO_ICCV21,TubeR_CVPR22,STMixer_CVPR23,EVAD_ICCV23}. The basic idea is to use a single video transformer to get features of all video frames, and then introduce learnable queries to mix with video features for actor localization and action recognition. Specifically, WOO~\cite{WOO_ICCV21} follows the Sparse RCNN~\cite{SparseRCNN_CVPR21} for localization, TubeR~\cite{TubeR_CVPR22} learns the action tubes following the classical DETR~\cite{DETR_ECCV20}, and STMixer~\cite{STMixer_CVPR23} follows the AdaMixer~\cite{AdaMixer_CVPR22} design that achieves the sate-of-the-art performance. The query-based design is advantageous in modeling the interaction between actors and actor context while simplifying the architecture as a single-stage design. However, none of these works could handle unseen actions in an open world. Therefore, we introduce an open-vocabulary action detection method in a query-based design to detect any actions.

\begin{figure*}[t]
    \centering
    \includegraphics[width=0.95\textwidth]{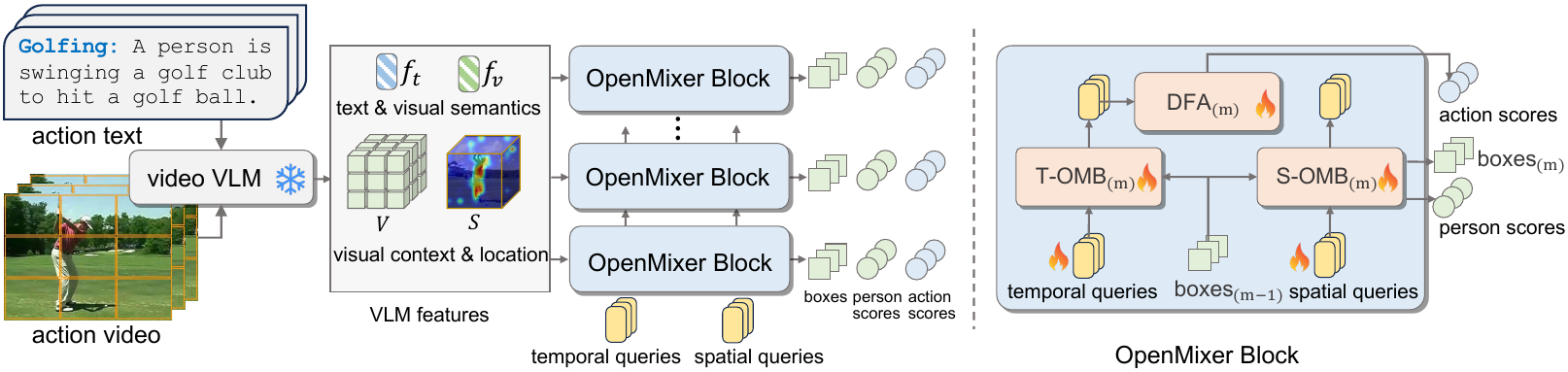}
    \caption{\textbf{Framework (left) and the OpenMixer Block (right).} Given a video and an open vocabulary of actions, we use prompted classes and a pre-trained video VLM to obtain all kinds of VLM features. 
    With a stack of cascaded OpenMixer blocks and spatial-temporal queries, the model predicts the action scores, person boxes, and their associated person scores for the OVAD task.}
    \label{fig:framework}
\end{figure*}

\textbf{Open-vocabulary Visual Understanding.} 
Thanks to the strong alignment capability of pre-trained visual language models (VLM), visual data from unseen classes in an open world can be recognized by the alignment between the visual feature and the text feature of the class names~\cite{OVSurvey_arXiv23,CLIP_ICML21}. This motivates a series of open-vocabulary works in object detection~\cite{OVRCNN_CVPR21,gdino,OVDETR_ECCV22,RegionCLIP_CVPR22,ROViT_CVPR23,VLDet_ICLR23,FVLM_ICLR23,CORA_CVPR23}, action recognition~\cite{ActionCLIP_arXiv21,XCLIP_ECCV22,EVL_ECCV22,ju2022prompting,ViFiCLIP_CVPR23,VitaCLIP_CVPR23}, and temporal action localization (OVTAL)~\cite{ju2022prompting,rathod2022open_bmvc22,nag2022zero,STAN_CVPR23}. For localizing unseen, the recent image-based open-vocabulary object detectors OV-DETR \cite{OVDETR_ECCV22} and CORA~\cite{CORA_CVPR23} share a common spirit with ours by injecting VLM semantics into the learnable queries. However, the query conditions in OV-DETR are class-specific such that they are not adaptive to test-time samples, and the two-stage training in CORA limits its flexibility in video domains. For the video understanding, the recent work~\cite{ju2022prompting},~\cite{rathod2022open_bmvc22}, and STAN~\cite{STAN_CVPR23} are built on existing image-based CLIP model. However, compared to the OVTAL task, the proposed OVAD task in this paper is even more challenging as it needs to distinguish between any actions in both spatial regions and temporal segments. In literature, the iCLIP~\cite{iCLIP_ICCVW23} aims for a zero-shot action detection, which does not consider seen actions in testing. Moreover, it skipped learning to localize actions by off-the-shelf person detectors~\cite{ren2015faster} and only learns to recognize the unseen actions, which lacks the adaptability to localize action-relevant persons. We noticed a concurrent work~\cite{wu2024open} for the OVAD problem, but it is two-stage designed and relies on extra large-scale region-text pre-training data, without fully exploiting the inherent detection knowledge of video-based VLMs (see~\cref{supsec:comp} for more comparative discussions). To the best of our knowledge, the OpenMixer is the first query-based OVAD model that can be combined with any video VLMs without region-level pre-training.

\section{Method}

In contrast to the closed-set video action detection~\cite{ACRN_ECCV18,ACTDet_ICCV17}, open-vocabulary action detection (OVAD) aims to recognize and spatiotemporally localize any human actions in videos, including action categories seen and unseen in training. Concretely, an OVAD model is learned from a training set of $N_{train}$ samples $\{(\mathbf{X}, \mathbf{Y})_i\rvert i=1,\dots,N_{train}\}$ where $\mathbf{X}$ denotes the training video and $\mathbf{Y}$ denotes the bounding box annotations on the keyframe that consists of box coordinates $\mathbf{b}$ and action category $y$. In training, an action $y$ is drawn from a fixed set of base action categories $\mathcal{C}_B$. In testing, the learned action detector could detect ``any'' actions in a given video from the open vocabulary $\mathcal{C}_B\cup\mathcal{C}_N$, where $\mathcal{C}_N$ contains any novel action categories.

\subsection{OpenMixer}

\begin{figure*}[t]
    \centering
    \subcaptionbox{\textbf{S-OMB} \label{fig:somb}}{
        \includegraphics[height=1.5in]{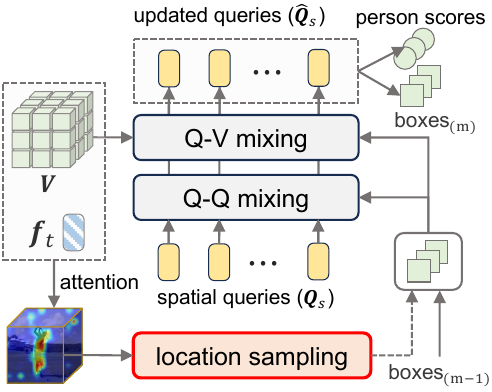}
    }
    \hspace{5mm}
    \subcaptionbox{\textbf{T-OMB}\label{fig:tomb}}{
        \includegraphics[height=1.5in]{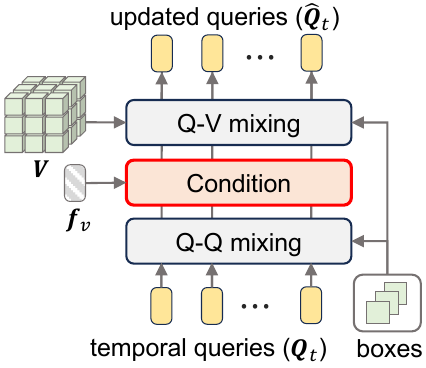}
    }
    \hspace{5mm}
    \subcaptionbox{\textbf{DFA}\label{fig:dfa}}{
        \includegraphics[height=1.5in]{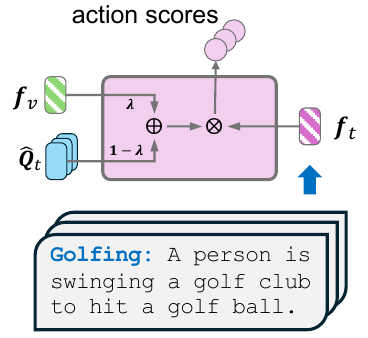}
    }
    \caption{\textbf{Spatial and Temporal OMB, and DFA.} In~\cref{fig:somb,fig:tomb}, the Q-Q and Q-V mixing modules aim to mix information among queries and across query-visual features, respectively. S-OMB is in Sec.~\ref{sec:S-OMB} where the dashed arrow is only used at the 1st stage. T-OMB is in Sec.~\ref{sec:TOMB} and DFA is in Sec.~\ref{sec:dfa}.}
    \vspace{-4pt}
\end{figure*}

In this paper, we propose the OpenMixer to solve the OVAD task. The OpenMixer model is developed within the family of query-based action DEtection TRansformers (DETR) \cite{STMixer_CVPR23,WOO_ICCV21,TubeR_CVPR22}. Basically, DETR-style models treat the action detection task as a set-to-set prediction problem, \ie, learning a sparse set of query features from videos to match with the ground truth boxes and action classes. Specific to the OVAD task, the action classes are predicted from an open vocabulary that contains both the base and novel actions. 

The OpenMixer is shown in Fig.~\ref{fig:framework} (left), given a video $\mathbf{X}$ and a list of text prompted action class as input, we leverage the visual and text encoders $\Psi_{\text{VE}}$ and $\Psi_{\text{TE}}$ of a pre-trained video VLM to obtain all features of the video and action text, \ie, $\mathbf{V},\mathbf{f}_v,\mathbf{S}=\Psi_{\text{VE}}(\mathbf{X})$ and $\mathbf{f}_t=\Psi_{\text{TE}}(y)$. Here, $\mathbf{V}$, $\mathbf{f}_v$ and $\mathbf{S}$ are the 4D patch-level video feature, video-level feature, and video attention, respectively, and $\mathbf{f}_t$ is the text feature of class $y$. Then, we build $M$ cascaded OpenMixer Blocks (\textbf{OMB}) to learn a set of $N$ spatial queries $\mathbf{Q}_s$ and $N$ temporal queries $\mathbf{Q}_t$ from $(\mathbf{V},\mathbf{S},\mathbf{f}_v,\mathbf{f}_t)$ for person detection and action classification, respectively.  The OMB takes as input all the features from VLM and the $\mathbf{Q}_s$ and $\mathbf{Q}_t$ to predict person boxes, person scores, and action scores. 

For the $m$-th OMB, as shown in Fig.~\ref{fig:framework} (right), it consists of a Temporal OpenMixer Block (T-OMB) $\Psi_{\alpha}$, a Spatial OpenMixer Block (S-OMB) $\Psi_{\theta}$, and a dynamically fused alignment (DFA). The S-OMB consists of prior location sampling, query-query (Q-Q) mixing by self-attention~\cite{Transformer_NIPS17} and query-video (Q-V) mixing by AdaMixer~\cite{AdaMixer_CVPR22}, while the T-OMB sequentially consists of Q-Q mixing, query conditioning, and Q-V mixing (see Fig.~\ref{fig:somb} and~\ref{fig:tomb} for reference). The DFA module recursively updates the $\mathbf{Q}_s$, $\mathbf{Q}_t$, and person boxes from the $(m\!-\!1)$-th OMB, and predict person scores and action scores. These three modules are developed for the OVAD task with the consideration of VLM \emph{semantics} and \emph{localizability}, which will be introduced in the following sections.

\subsection{Localizability Prior for Spatial OMB}
\label{sec:S-OMB}

A major challenge for one-stage query-based detectors is the low convergence of localization. One of the causes is the lack of prior knowledge of object locations. Specific to the action detection, recent two-stage action detectors~\cite{ConstCL_CVPR22,iCLIP_ICCVW23,HIT_WACV23,EVAD_ICCV23} address location prior by an off-the-shelf person detector and RoIAlign~\cite{maskrcnn} cropping, but the feature cropping lacks the spatiotemporal context and suffer from representation gap when a pre-trained VLM is introduced. For recent query-based action detectors~\cite{WOO_ICCV21,TubeR_CVPR22,STMixer_CVPR23}, the prior knowledge of the person locations is missing in their design. Therefore, when it comes to the OVAD task by VLMs, a natural question is that, \emph{Can we obtain the prior locations of actors from pre-trained VLMs in a cheap way?} Motivated by these considerations, we resort to the visual attention from a pre-trained VLM.

\vspace{1mm}
\noindent\textbf{Prior Locations from VLM Attention}. Visual attention maps are traditionally represented by the class activation map (CAM) to visually explain recognition models~\cite{CAM_CVPR16,GradCAM_ICCV17}. In the era of ViT~\cite{ViT_ICLR21} and VLM~\cite{CLIP_ICML21}, recent works~\cite{DINO_MHSA,HilaCAM_ICCV21,CLIPSurgery} propose to use the multi-head self-attention (MHSA) of the last ViT layer, or the gradient-weighted accumulative product over multi-layer self-attention. However, MHSA is not visually faithful due to the high redundancy of video tokens, and the gradient-based methods suffer from a huge computational cost on video VLMs and ad-hoc implementation for different VLMs. Moreover, due to the lack the token-level video-text correlation, their attention map does not closely relevant to the action specified by the vocabulary. Therefore, an efficient and structure-agnostic CAM is preferable to large video VLMs, which motivates us to use patch-text correlation as VLM attention to encode the location priors. 

Specifically, with the $D$-dimensional 4D video feature $\mathbf{V}\in\mathbb{R}^{T\times hw\times D}$ where $T$ is the number of frames and $hw$ is the number of visual tokens in each frame, the holistic video feature $\mathbf{f}_v\in \mathbb{R}^D$, and the text features of $C$ classes as $\mathcal{F}_t=[\mathbf{f}_t^{(1)},\ldots,\mathbf{f}_t^{(C)}]^\top$. The features are $L_2$ normalized. We first get the pre-matched text feature $\mathbf{f}_t$ by maximum similarity: $\mathbf{f}_t=\arg\max_{\mathbf{f}_t\in\mathcal{F}_t} \mathbf{f}_v^{\top}\otimes \mathbf{f}_t$, since we do not have access to the class label in testing. Thus, the inner-product between $\mathbf{f}_t$ and $\mathbf{V}$ determines the patch-text correlation: $\mathbf{S} = \mathbf{V}\otimes \mathbf{f}_t$. Furthermore, as discussed in~\cite{RITSM,CLIPSurgery}, the q-v attention in self-attention layers shows an opposite heatmap where the foreground regions are associated with low attention value. In practice, we also observed this issue so that similar to~\cite{RITSM}, our CAM is determined by the reversed patch-text similarity: $\hat{\mathbf{S}} = 1 - \mathbf{S}$ (see detailed explanation in~\cref{appd:reverse_attn}). By reshaping and spatial interpolation over $\hat{\mathbf{S}}$, the attention map is obtained for prior location sampling. We treat the $\hat{\mathbf{S}}$ as the prior distribution of person locations indicated by the VLM, thus the top-$N$ positions are sampled as the initial boxes centers: $\{(u,v)_i|i=1,\ldots,N\}\!\sim\!\hat{\mathbf{S}}(u,v,k)$ where $(u,v)$ are 2D coordinates on the keyframe $k$ and $N$ is the number of queries.

\vspace{1mm}
\noindent\textbf{Spatial OMB.}
With the sampled prior locations, the S-OMB (see Fig.~\ref{fig:somb}) that consists of Q-Q and Q-V mixing modules takes as input the video patch features $\mathbf{V}$ and the box prediction $\hat{\mathbf{b}}_{m-1}$ of the previous $(m\!-\!1)$-th stage, to update the spatial queries by $\hat{\mathbf{Q}}_s = \Psi_{\theta_m}(\mathbf{V},\mathbf{Q}_s,\hat{\mathbf{b}}_{m-1})$. The updated spatial queries $\hat{\mathbf{Q}}_s$ are used to predict the person scores $\hat{\mathbf{o}}_m$ and person box offsets $\Delta\hat{\mathbf{b}}_m$ by MLP. Then, the predicted boxes at stage $m$ are updated by $\hat{\mathbf{b}}_{m} = \hat{\mathbf{b}}_{m-1} + \Delta\hat{\mathbf{b}}_{m}$, where initial box queries $\hat{\mathbf{b}}_0$ consist of the sampled prior locations and the video spatial range.

The technical intuition behind the design is to encourage the proposed Spatial OMB to learn the box offset $\Delta\mathbf{b}$ starting from the prior locations inherited from the pre-trained VLM. Besides, compared to~\cite{STMixer_CVPR23} that uses the fixed non-informative frame centers as prior locations, our VLM attention-based prior locations are adaptive to the test-time video content and vocabulary, which improves not only the seen action localization but also the generalization to the unseen (see~\cref{tab:main_res} ZSR+TL section).

\subsection{Adaptive Semantics for Temporal OMB}
\label{sec:TOMB}

For the query-based OVAD models, temporal queries are expected to be discriminative for both base and novel actions. This requires a strong capability of content decoding for the query-video (Q-V) mixing module. The pioneering work DETR~\cite{DETR_ECCV20} uses cross-attention while~\cite{AdaMixer_CVPR22,STMixer_CVPR23} adopt the MLP-Mixer~\cite{MLPMixer_NIPS21}. However, without VLM semantics, these approaches inevitably overfit the seen class data and are unable to detect the unseen. Recent works~\cite{OVDETR_ECCV22,CORA_CVPR23} rightly address the importance of VLM semantics for the query features, but they lack the adaptability to the test-time visual content due to the class-wise semantic condition in~\cite{OVDETR_ECCV22} and the region prompting in~\cite{CORA_CVPR23}. These motivate us to propose the Temporal OMB that exploits adaptive semantics from pre-trained VLMs.

\vspace{1mm}
\noindent\textbf{Temporal OMB.}
As dipicted in Fig.~\ref{fig:tomb}, with the temporal queries $\mathbf{Q}_t$ and the predicted boxes $\hat{\mathbf{b}}_m$ at the current stage $m$, the queries are updated by interacting with the video features $\mathbf{V}$ and $\mathbf{f}_v$ by the function $\hat{\mathbf{Q}_t} = \Psi_{\alpha_m}(\mathbf{V},\mathbf{Q}_t, \mathbf{f}_v, \hat{\mathbf{b}}_{m})$. To achieve our motivation of using adaptive semantics, we propose a query update:
\begin{equation}
    \hat{\mathbf{Q}}_t = \Psi_{qv}\left(\Psi_{qq}(\mathbf{Q}_t,\mathbf{b}) \oplus \mathbf{f}_{v},\mathbf{V},\mathbf{b}\right),
\label{eq:pre_cond}
\end{equation}
where $\Psi_{qq}$ and $\Psi_{qv}$ are Q-Q mixing and Q-V mixing modules by self-attention~\cite{Transformer_NIPS17} and AdaMixer~\cite{AdaMixer_CVPR22}, respectively. Here, $\mathbf{f}_{v}$ is the adaptive semantic condition by the pre-trained VLM video feature, which is broadcastly added (denoted as $\oplus$) to the output of Q-Q mixing. 

\vspace{1mm}
\noindent\textbf{Remark.} Note that the adaptiveness of the semantic condition stems from the test-time video feature $\mathbf{f}_v$. Alternatively, when the semantic condition $\mathbf{f}_v$ is changed to $\mathbf{f}_t$ over $C$ classes, it is equivalent to the way in~\cite{OVDETR_ECCV22}. However, we empirically show this leads to inferior performance (see~\cref{tab:cond}) especially for the seen action detection. The inferiority can be attributed to the lack of adaptability to test-time video content. Besides, as another alternative, the post-condition that places the condition $\mathbf{f}_v$ after the Q-V mixing, \ie, $\hat{\mathbf{Q}}_t = \Psi_{qv}\left(\Psi_{qq}(\mathbf{Q}_t,\mathbf{b}),\mathbf{V},\mathbf{b}\right) \oplus \mathbf{f}_{v}$, the module $\Psi_{qv}$ is thus to learn the residual of $\mathbf{f}_v$. We empirically found that our pre-condition by Eq.~\ref{eq:pre_cond} is superior to the post-condition, potentially because of the better query features used to learn the important Q-V mixing module.

\subsection{Dynamically Fused Alignment} 
\label{sec:dfa}

To recognize both seen and unseen actions, the model needs to learn discriminative region-wise visual features to align with seen actions, while keeping the generalizable knowledge of the pre-trained VLMs to align with the unseen actions. Dealing with the two goals is challenging. A line of recent approaches uses model adaptation by prompt tuning~\cite{ju2022prompting,DetPro_CVPR22,nag2022zero,VitaCLIP_CVPR23,CORA_CVPR23,iCLIP_ICCVW23}, adapters~\cite{STAdapter_NIPS,CLIPAdapter_IJCV23}, and gradient preserving~\cite{OpenVCLIP_ICML23,ProGrad_ICCV23}. However, these methods either struggle in generalization to novel categories or need to back-propagate through the large VLM that incurs huge computational costs, especially for long videos. Therefore, we resort to a dynamically fused alignment (DFA) for open-vocabulary action recognition, which is lightweight in design and works well for both seen and unseen actions.

Specifically, as shown in~\cref{fig:framework,fig:dfa}, the DFA is formulated to learn the action classification at each stage $m$, \ie, $\hat{\mathbf{y}}_{m} = \Psi_{\boldsymbol{\lambda}_m}(\hat{\mathbf{Q}_t}, \mathbf{f}_v, \mathbf{f}_t)$, where $\hat{\mathbf{y}}_m$ are the predicted actions for all queries $\hat{\mathbf{Q}_t}$ and the $\boldsymbol{\lambda}_m$ are the learnable parameters. This module includes \emph{dynamic feature fusion} and \emph{query-text alignment}.

\vspace{1mm}
\noindent\textbf{Dynamic Feature Fusion.} This step aims to fuse the video-level feature $\mathbf{f}_v$ into each of the queries $\hat{\mathbf{Q}_t}$ dynamically. Specifically, we first repeat $N$ times of the $\mathbf{f}_v$ to be $\mathbf{F}_v\in\mathbb{R}^{N\times D}$. Then, the fusion between $\mathbf{F}_v$ and $\hat{\mathbf{Q}}_t$ is achieved by $\tilde{\mathbf{F}}_v = \boldsymbol{\lambda} \odot \mathbf{F}_v + (1 - \boldsymbol{\lambda}) \odot \hat{\mathbf{Q}}$, where $\boldsymbol{\lambda}\in\mathbb{R}^{N\times 1}$ are learnable in training. The intuition behind the query-specific learnable $\boldsymbol{\lambda}$ is that, it allows the dynamic contributions of the video-level knowledge from $\mathbf{f}_v$ to the different learnable queries in the set-matching training.

\input{tables/benchmark}

\vspace{1mm}
\noindent\textbf{Query-Text Alignment.} To make the classification decision by $\tilde{\mathbf{F}}_v$ and open vocabulary of actions, for the action category, we leverage GPT-4~\cite{GPT_arXiv2023} to generate multiple visually descriptive action prompts for each category (see the prompts in~\cref{supsec:prompts}). With VLM text encoder, the aggregated text features of $\mathcal{C}$ classes are represented as $\mathbf{F}_t\in\mathbb{R}^{\mathcal{C}\times D}$, where $\mathcal{C}$ is the number of classes. Eventually, we use the softmax of visual-text cosine similarity to represent the multi-class classification probability: $P(\hat{y}|\hat{\mathbf{Q}})=\text{softmax}(\tilde{\mathbf{F}}_v\otimes \mathbf{F}_t^\top/\tau)$ where $\tau$ is the VLM temperature.  In testing, the open-vocabulary action recognition for all queries is achieved by finding the maximum visual-text cosine similarity: $\hat{y}=\arg\max_{y\in\mathcal{C}} (\tilde{\mathbf{F}}_v\otimes \mathbf{F}_t^\top)$. 

Note that we do not include the spatial queries $\mathbf{Q}_s$ as the input of our DFA module. This makes the T-OMB $\Psi_{\alpha}$ and the S-OMB $\Psi_{\theta}$ to be decoupled in training such that the person localization is class-agnostic, which is essential for open-vocabulary tasks according to~\cite{nag2022zero,CORA_CVPR23}. 

\subsection{Training and Inference}
In training, for action localization, we adopt the regular set matching loss following the DETR literature~\cite{DETR_ECCV20,SparseDETR_ICLR22,AdaMixer_CVPR22}: $\mathcal{L}_{set} = \mathcal{L}_{bce} + \mathcal{L}_{L_1} + \mathcal{L}_{giou}$, where $\mathcal{L}_{bce}$ is a binary cross-entropy loss for person score prediction, $\mathcal{L}_{L_1}$ and $\mathcal{L}_{giou}$ are the coordinate distance and GIoU distance~\cite{GIOU_CVPR19} between predicted and ground truth boxes, respectively. Then, we use the Hungarian matching~\cite{DETR_ECCV20} to find the optimal bipartite matching between the predicted and ground truth boxes for each video. For action recognition, we use a multi-class cross-entropy loss $\mathcal{L}_{act}$ so that the total loss for training is $\mathcal{L}_{total}=w_1\mathcal{L}_{set}+w_2\mathcal{L}_{act}$ where the hyperparameters $w_1$ and $w_2$ are used to balance between the two subtasks.

During inference, the thresholded person scores determine the kept person boxes, while the action scores assign the action categories to boxes from input class categories.

\section{Experiments}

\vspace{1mm}
\noindent\textbf{Datasets.} Our method is implemented on two commonly-used action detection datasets, \ie, J-HMDB~\cite{jhmdb} and UCF101-24~\cite{ucf101}. J-HMDB dataset contains per-frame annotated bounding boxes of persons along with 21 action classes. Similar to~\cite{ju2022prompting,iCLIP_ICCVW23}, with 50\% of actions as the novel classes, we randomly split it into 10 base classes for training and 11 novel classes for testing, which results in 10,570 training samples and 9,139 testing samples. UCF101-24 dataset is a subset of UCF101~\cite{ucf101}. It is also per-frame annotated for action detection and contains 24 action classes. With the same 50\% splitting strategy, we split it into 12 base classes for training and 12 novel classes for testing. Similar to~\cite{ju2022prompting,nag2022zero}, we also report results on other random splits with ratios of both 50\%-50\% and 75\%-25\% in~\cref{supsec:addres}. We will release all data splits. 

\vspace{1mm}
\noindent\textbf{Evaluation criteria.} Following the standard paradigm in action detection literature~\cite{yowo19,HIT_WACV23,WOO_ICCV21,TubeR_CVPR22,STMixer_CVPR23}, the model performance is evaluated by video mAP. It evaluates the spatiotemporal action tubes of the detected bounding boxes over the classification and 3D intersection-over-union (IoU). Following~\cite{HIT_WACV23}, the 3DIoU threshold is set to 0.5 for J-HMDB and 0.2 for UCF101-24, respectively.

\vspace{1mm}
\noindent\textbf{Implementation details.} We experiment with two VLMs including the image pre-trained OpenAI CLIP~\cite{CLIP_ICML21} and video pre-trained CLIP-ViP~\cite{CLIPViP_ICLR2023}. We use the same ViT-B/16 architecture for the two VLMs. The VLMs are kept frozen in training. For the image CLIP, we get video-level semantic features by temporal mean pooling. We obtain the patch token features of the last ViT layer and use them to construct the 4D pyramid feature $\mathbf{V}$ by multi-scale residual convolutions. By default, we set the number of queries and OMB stages to 100 and 3, respectively. In training, we set the mini-batch size to 16 and frame sampling by $16\times 1$. The weight of the set loss $\mathcal{L}_{set}$ and action loss $\mathcal{L}_{act}$ are set to 2.0 and 48.0, respectively. Following~\cite{DETR_ECCV20,WOO_ICCV21,STMixer_CVPR23}, each intermediate stage is individually supervised by the loss $\mathcal{L}_{set}$ and $\mathcal{L}_{act}$. We set the base learning rate to 1e-5 and use the AdamW~\cite{AdamW_ICLR18} optimizer to train models for 12 epochs on 4 RTX 6000Ada or 2 A100 (80G) GPUs. In testing, the person detection threshold is set to 0.6. We individually test the base and novel classes and report their video-mAP results and the mean on all categories. In~\cref{supsec:addres}, we also report generalized zero-shot testing by giving a complete list of base and novel classes. Our model inference speed is 0.23 s/video per A6000 GPU, with 587M parameters based on CLIP-ViP/B16 VLM. Other details are in~\cref{supsec:impl}.

\vspace{1mm}
\noindent\textbf{OVAD task settings.} To benchmark methods on OVAD task, three settings are presented considering if the localization and classification are trained or not. 
\begin{itemize}
\vspace{-2mm}
\itemsep0em
    \item \textbf{ZSR+ZSL} (zero-shot action recognition and actor localization): without any training, we only use pre-trained person detectors such as Mask RCNN~\cite{maskrcnn} to detect persons, and use pre-trained video VLMs such as CLIP-ViP~\cite{CLIPViP_ICLR2023} for open-vocabulary recognition.
    \item \textbf{ZSR+TL} (zero-shot action recognition and trainable actor localization): we use pre-trained CLIP-ViP~\cite{CLIPViP_ICLR2023} to perform video-level action recognition while training the localization modules to detect persons. 
    \item \textbf{E2E} (end-to-end learning): we train and test models in an end-to-end way by using raw videos and vocabulary as input. In this setting, we compare with STMixer~\cite{STMixer_CVPR23} with the same CLIP-ViP~\cite{CLIPViP_ICLR2023} backbone and investigate the soft prompt by CoOp~\cite{CoOp_IJCV22}. 
\end{itemize}

\subsection{Comparative Results}
 
The main results are reported in~\cref{tab:main_res}. To analyze the baseline performance, we summarize the discussion below.

\vspace{1mm}
\noindent\textbf{Zero-shot recognition and localization.} In the ZSR+ZSL setting, the findings are as follows. First, region-level features (the 1st row) by RoIAlign~\cite{maskrcnn} perform significantly worse than the video-level features (the 3rd row). This indicates that the RoI-cropped features from VLM suffers from a large representation gap between the video-level pre-training and downstream region-level recognition. Second, the descriptive \textbf{GPT}-generated prompts (the 3rd row) achieve a better performance than the handcrafted (\textbf{HC}) prompt such as the ``\texttt{a video of person [CLS]}" (the 2nd row). This can be explained by the more transferable knowledge in the GPT prompts than the handcraft ones. 

\vspace{1mm}
\noindent\textbf{Zero-shot recognition with learnable localization.} Under the ZSR+TL setting, we observe a significant superiority of Spatial OpenMixer to the STMixer baseline, with more than 10\% performance gain on J-HMDB dataset. Since the training is only encouraged to localize actors in videos, the outperformance suggests a good exploitation of the localizability in pre-trained VLMs.

\input{tables/iclip}
\input{tables/ablation}

\vspace{1mm}
\noindent\textbf{End-to-end learnable OVAD.} For the E2E setting, the OpenMixer (the last row) outperforms the simple STMixer baseline (STMixer+VLM) by large margins, with $7.69\%$ and $54.89\%$ of video mAP gains on base and novel categories of the J-HMDB dataset, respectively. Besides, we explored the widely-used VLM adaptation method CoOp~\cite{CoOp_IJCV22} that optimizes the context of class names, \ie, prompt tuning. From~\cref{tab:main_res}, we observe that CoOp improves the base class performance with sacrifice on the novel classes, while the GPT-prompted OpenMixer achieves much better performance on novel classes. Lastly, we notice the relatively smaller numbers on UCF101-24 than those on J-HMDB. This reflects the challenging aspects of the UCF101-24 dataset such as the long duration ($\sim10\times$ longer), heavy background bias, and multi-person scenarios.

\vspace{1mm}
\noindent\textbf{Zero-shot action detection.} We note the iCLIP~\cite{iCLIP_ICCVW23} defines the zero-shot action detection (ZSAD) task which is different from our OVAD task. The ZSAD only cares about the samples from novel classes while OVAD values both the base and novel classes. Therefore, ZSAD uses all samples from base classes in training and only tests on novel classes. Following the same settings as iCLIP, the results in~\cref{tab:iap_splits} show that our method could achieve much better performance than iCLIP, even though iCLIP relies on pre-detected person boxes from YOWO~\cite{yowo19}.

\input{tables/condition}
\input{tables/fusion}

\subsection{Ablation Study}

\input{figures/plots}

\input{figures/visualize}

In this section, we analyze the properties of the OpenMixer model on the J-HMDB dataset. Results of the component-wise ablation are reported in~\cref{tab:ablation}. It shows that all three components could work well. Specifically, without S-OMB which means the attentional location prior is removed, the performance drops significantly especially for the novel classes. If the DFA is removed, we only use the pre-trained VLM feature for zero-shot recognition, it shows that the base class performance is the worst. Without T-OMB which means the semantic condition is removed and both spatial and temporal queries are used for recognition, it shows a decrease of $4.74$\% and $1.15$\% on base and novel actions, respectively. 

\vspace{1mm}
\noindent\textbf{Query condition strategies.} Specific to the T-OMB, we further investigate different alternatives of query condition strategies in~\cref{tab:cond}, which shows the following observations: (1) Without any condition, it performs worse on the base class with $4.89$\% mAP drop. (2) Pre-condition performs much better than post-condition on base classes ($+2$\%), with negligible performance drop on the novel classes ($-0.19$\%). This can be explained that the pre-condition alleviates the difficulty of content decoding by the following Q-V mixing module. (3) The additional condition on spatial queries (SQ) hurts the performance on both the base and novel classes, because this essentially makes the recognition and localization entangled in training. (4) When using text feature $\mathbf{f}_t$ as a condition, base class performance significantly decreased ($-20.50$\%) and novel class performance also decreased a bit. This is due to the large semantic gap between text feature $\mathbf{f}_t$ and patch-wise video token features $\mathbf{V}$, suggesting that the test-time adaptive $\mathbf{f}_v$ is preferable even though $\mathbf{f}_v$ and $\mathbf{f}_t$ are semantically aligned.

\vspace{1mm}
\noindent\textbf{Feature fusion strategies.} To validate the design choice of our DFA module, we explored different feature fusion strategies, as shown in~\cref{tab:fusion}. The results show that only using the learned query feature $\hat{\mathbf{Q}}_t$ (the 1st row) performs much worse performance on the novel classes, indicating the loss of generalization. If only using the pre-trained feature $\mathbf{F}_v$ (the 2nd row), the model cannot work well on the base classes which indicates an under-fitting to the task. If fusing the features by simple averaging, the performance still lags behind ours as it is not adaptive to the variety in queries. Moreover, we notice that~\cite{STMixer_CVPR23} uses both spatial and temporal queries for recognition by MLP layers. Thus, we additionally include the spatial queries $\hat{\mathbf{Q}}_s$ by concatenation with temporal queries $\hat{\mathbf{Q}}_t$ and use MLP layers for dimension reduction. We observe the performance drop, which can be explained as the MLP layers eliminate the benefits of the semantic conditions and makes localization and recognition entangled in training.

\vspace{1mm}
\noindent\textbf{Number of queries and OMB stages.} In~\cref{fig:nquery} and~\ref{fig:nstage}, we show that using 100 queries and 3 OMB stages achieves the best average mAP. The figures also indicate that the number of OMB stages is more important than the number of queries, as the bipartite matching could handle the redundant queries in training. The decreasing trend with more than three OMB stages can be attributed to the risk of overfitting to training data. \emph{More interesting results and discussions can be found in\textbf{~\cref{supsec:addres}}}.

\vspace{1mm}
\noindent\textbf{Qualitative results.} We visualize results on the J-HMDB novel categories in~\cref{fig:vis}. They show that OpenMixer could precisely localize and confidently recognize those unseen actions, even though there are multiple persons. More visualizations are in~\cref{supsec:vis}.

\vspace{1mm}
\noindent\textbf{Limitations and Future Work.} 
The recent large-scale action detection dataset AVA~\cite{gu2018ava} is not included in this paper, as we emphasize on the adaptation of existing pre-trained VLMs for downstream small datasets. In the future, similar to the concurrent work~\cite{wu2024open}, we will explore how to effective pre-train on AVA to benefit for a more general audience.

\section{Conclusion}

We present an open-vocabulary action detection method OpenMixer to detect any human actions in videos. It is a query-based detection transformer that fully exploits the semantics and localizability of pre-trained VLMs. Furthermore, we build OVAD benchmarks that extensively evaluate baselines and our model under various settings, showing the superiority of the OpenMixer.

\vspace{1mm}
\noindent\textbf{Acknowledgement.} Wentao Bao and Yu Kong are partially supported by the Army Research Office (ARO) grant W911NF-24-1-0385 and Office of Naval Research (ONR) grant N00014-23-1-2046. Any opinions, findings, and conclusions or recommendations expressed in this material are those of the authors and do not necessarily reflect the views of ARO nor ONR.

{\small
\bibliographystyle{ieee_fullname}
\bibliography{main,supp}
}

\clearpage
\input{arXiv_supp}

\end{document}

%% file: tables/benchmark.tex
\begin{table*}[t]
    \centering
    \small
    \vspace{-2mm}
    \setlength{\tabcolsep}{5.2mm}
    \begin{tabular}{l|l|ccc|ccc}
         \toprule
         \multirow{2}{*}{Settings} & \hfil\multirow{2}{*}{Models}\hfill & \multicolumn{3}{c|}{J-HMDB} & \multicolumn{3}{c}{UCF101-24} \\
         \cline{3-8}
         & & Mean & Base & Novel   & Mean & Base & Novel \\
         \hline
         \multirow{3}{*}{\shortstack[c]{ZSR+ZSL}} & Region + GPT & 31.86  & 30.06  & 33.51   & 19.92  & 21.54  & 18.29  \\ 
         & Video + HC & 54.40  & 49.89  & 58.51   & 31.04  & 31.43  & 30.64   \\ 
         & Video + GPT & 66.73  & 64.61  & 68.66  & 35.01  & 34.59  & 35.43   \\ 
         \midrule
         \multirow{2}{*}{\shortstack[c]{ZSR+TL}} & STMixer~\cite{STMixer_CVPR23}  & 63.53  & 58.27  & 68.31   & 36.66  & 45.26  & 28.07  \\
         & \textbf{Spatial OpenMixer} & 74.06  & 68.04  & 79.53   & 40.32  & 48.80  & 31.85    \\
         \midrule
         \multirow{4}{*}{E2E} 
         & STMixer~\cite{STMixer_CVPR23} & 49.16  & 73.06  & 27.44 & 33.72  & 60.91  & 6.54    \\
         & STMixer~\cite{STMixer_CVPR23} (\scriptsize{w. CoOp~\cite{CoOp_IJCV22}})  & 42.27   & 75.66   & 11.91   & 36.12   & 60.42   & 11.81   \\
         & \textbf{OpenMixer} (\scriptsize{w. CoOp~\cite{CoOp_IJCV22}}) & 86.86   & 94.18   &  80.20    &  45.11  & 62.48  & 27.75  \\
         & \textbf{OpenMixer} & {86.34}  & {90.75}  & {82.33}   & 47.71   & 61.18   & 34.23   \\
    \bottomrule
    \end{tabular}
    \caption{\textbf{OVAD results.} Note that all methods including STMixer~\cite{STMixer_CVPR23} use the same pre-trained CLIP-ViP~\cite{CLIPViP_ICLR2023} as the frozen VLM and are evaluated by video mAP. For the ZSR+ZSL setting, we use off-the-shelf Mask RCNN~\cite{maskrcnn} as the ZSL person localizer and use either handcrafted (\textbf{HC}) or GPT-generated (\textbf{GPT}) prompts for either video- or region-level zero-shot recognition.}
    \label{tab:main_res}
\end{table*}

%% file: tables/iclip.tex
\begin{table}[t]
    \centering
    \small 
    \vspace{-2mm}
    \setlength{\tabcolsep}{8mm}
    \begin{tabular}{l|cc}
         \toprule
         Models & f-mAP  & v-mAP  \\
         \hline
         iCLIP~\cite{iCLIP_ICCVW23} & 65.41   & --  \\ 
         OpenMixer & \textbf{77.06}  & \textbf{81.20}  \\
         \bottomrule
    \end{tabular}
    \caption{\textbf{Zero-shot action detection}. Following the same 75\%-25\% split as ~\cite{iCLIP_ICCVW23}, we report both the frame- and video-level mAP (f-mAP and v-mAP) on novel classes of J-HMDB. 
    }
    \label{tab:iap_splits}
\end{table}

%% file: tables/ablation.tex
\begin{table}[t]
    \centering
    \small 
    \vspace{-2mm}
    \setlength{\tabcolsep}{2.7mm}
    \begin{tabular}{ccc|ccc}
         \toprule
         S-OMB & DFA  & T-OMB & Mean & Base & Novel  \\
         \hline
         \xmark  & \checkmark & \checkmark & 81.77  & 86.32  & 77.64  \\ 
         \checkmark  & \xmark & \checkmark & 74.06   & 68.04    & 79.53  \\ 
         \checkmark & \checkmark & \xmark & 83.47 & 86.01 & 81.18   \\
         \checkmark & \checkmark & \checkmark  & \textbf{86.34}  & \textbf{90.75}  & \textbf{82.33}  \\
    \bottomrule
    \end{tabular}
    \caption{\textbf{Ablation study} of each proposed component.}
    \label{tab:ablation}
\end{table}

%% file: tables/condition.tex
\begin{table}[t]
    \centering
    \small
    \vspace{-2mm}
    \setlength{\tabcolsep}{2.2mm}
    \begin{tabular}{lcc|ccc}
         \toprule
         Methods & Queries & Modalities & Mean & Base  & Novel \\
         \midrule
         \multicolumn{3}{c|}{w/o condition}   & 83.99  & 85.86  & 82.28  \\
         \midrule
         post \hfill & TQ & video $\mathbf{f}_v$ & 85.48 & 88.74 & \textbf{82.52}  \\
         \cmidrule{1-3}
         \multirow{3}{*}{pre}\hfill  & TQ, SQ & video $\mathbf{f}_v$ & {85.66} & {90.29} & 81.45   \\
         \cmidrule{2-3}
             & \multirow{2}{*}{TQ} & text $\mathbf{f}_t$ & 76.36 & 70.25 & 81.92   \\
             \cmidrule{3-3}
             &  & video $\mathbf{f}_v$  & \textbf{86.34} & \textbf{90.75} & {82.33}   \\
         \bottomrule
    \end{tabular}
    \caption{\textbf{Results of query conditions.} The post/pre and TQ/SQ means that the conditional feature (from video $\mathbf{f}_v$ or from text $\mathbf{f}_t$) is placed after/before the Q-V mixing on the temporal queries (TQ) or spatial queries (SQ).}
    \label{tab:cond}
\end{table}

%% file: tables/fusion.tex
\begin{table}[t]
    \centering
    \small
    \vspace{-2mm}
    \setlength{\tabcolsep}{3.5mm}
    \begin{tabular}{l|ccc}
         \toprule
         Methods  & Mean & Base  & Novel\\
         \midrule
         w/o. $\mathbf{F}_v$ ($\boldsymbol{\lambda}=0$) & 68.84  & 88.94  & 50.58 \\  
         w/o. $\hat{\mathbf{Q}}_t$ ($\boldsymbol{\lambda}=1$) & 74.06   & 68.04    & 79.53 \\
         w/o. dynamics ($\boldsymbol{\lambda}=0.5$)  & 51.48  & 63.08  & 40.93   \\  
         concat $[\hat{\mathbf{Q}}_s;\hat{\mathbf{Q}}_t]$ \& mlp   & 85.51  & 89.19  & 82.17  \\ 
         Ours  & \textbf{86.34}  & \textbf{90.75}  & \textbf{82.33}   \\
         \bottomrule
    \end{tabular}
    \caption{\textbf{Results of fusion strategies.} We explored different strategies to fuse the pre-trained $\mathbf{F}_v$ and learnable queries ($\hat{\mathbf{Q}}_t$ or $\hat{\mathbf{Q}}_s$) within our DFA module.}
    \label{tab:fusion}
\end{table}

%% file: figures/plots.tex
\begin{figure}[t]
    \centering
    \subcaptionbox{number of queries \label{fig:nquery}}{
        \includegraphics[width=0.45\linewidth]{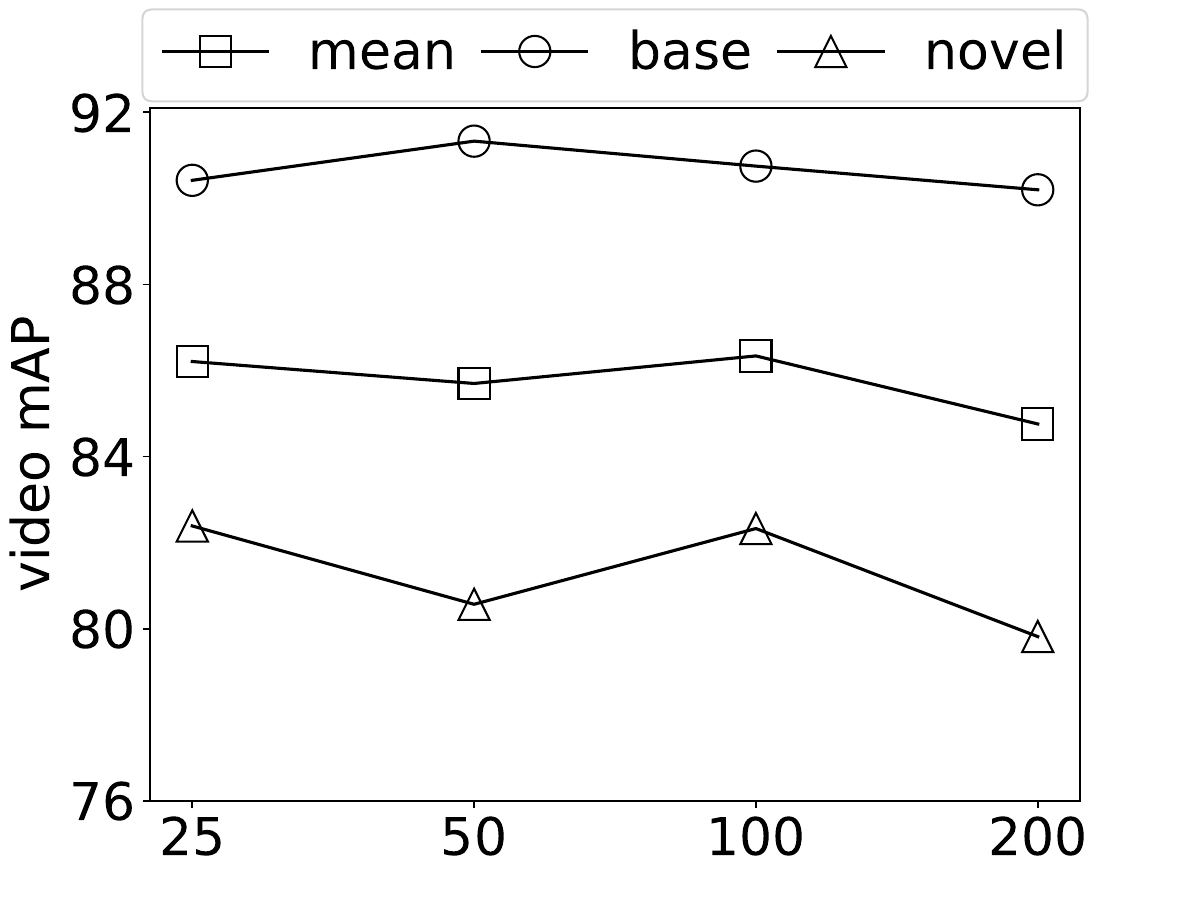}
    }
    \hfill
    \subcaptionbox{number of stages\label{fig:nstage}}{
        \includegraphics[width=0.45\linewidth]{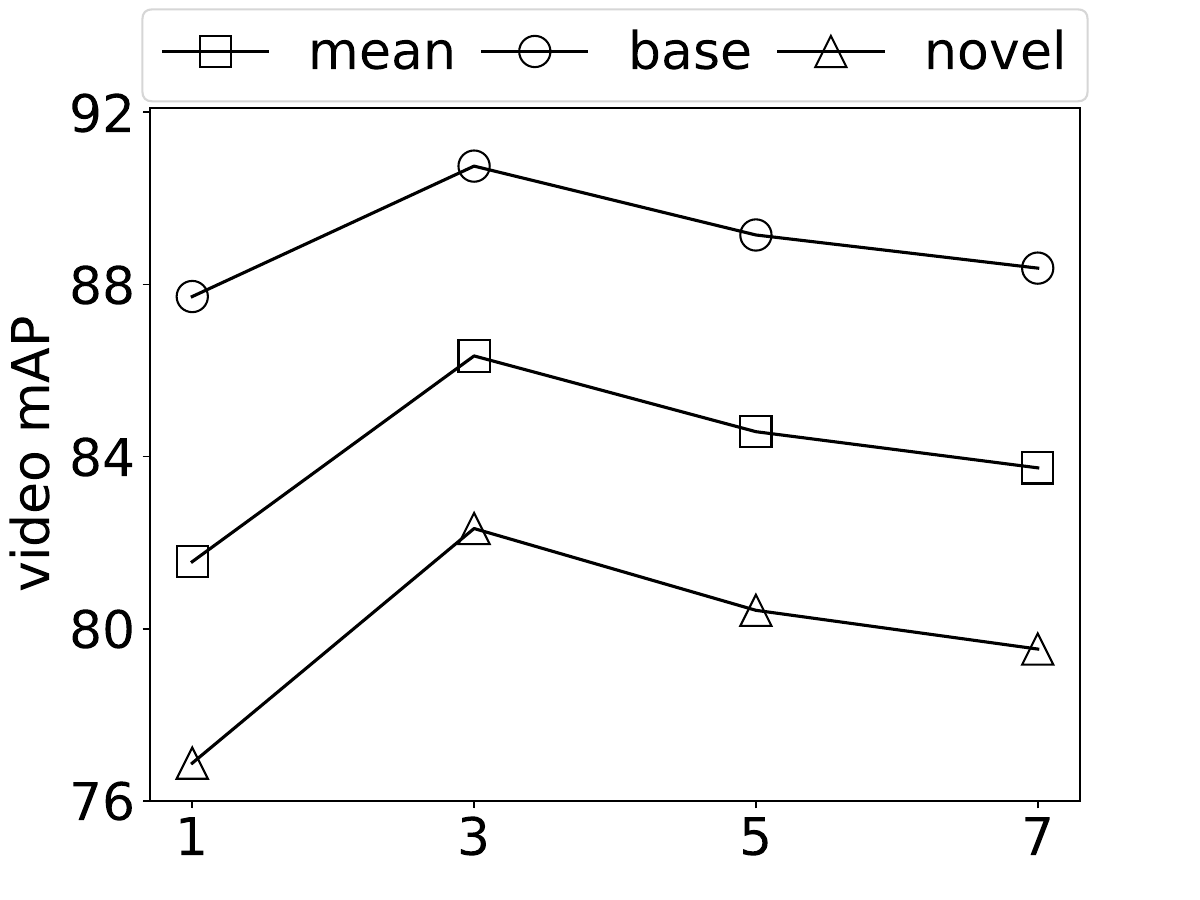}
    }
    \vspace{-2mm}
    \caption{\small{\textbf{Hyperparameters.} We show the video mAP with respect to different numbers of learnable queries and OMB stages.}}
    \vspace{-5pt}
\end{figure}

%% file: figures/visualize.tex
\newcommand{\framewidth}{0.16\linewidth}

\begin{figure*}[t]
\footnotesize
\centering
\renewcommand{\tabcolsep}{0.7pt} %
\begin{tabular}{ccccccc}
& $t/T=1/6$ & $t/T=2/6$ & $t/T=3/6$ & $t/T=4/6$ & $t/T=5/6$ & $t/T=1$
\\
\parbox[c]{4mm}{\multirow{1}{*}[3em]{\rotatebox[origin=c]{90}{kick ball}}} &
\includegraphics[width=\framewidth]{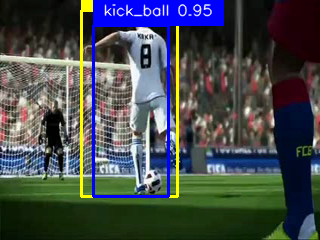} &
\includegraphics[width=\framewidth]{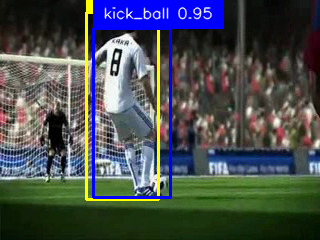} &
\includegraphics[width=\framewidth]{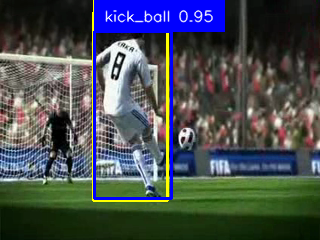} &
\includegraphics[width=\framewidth]{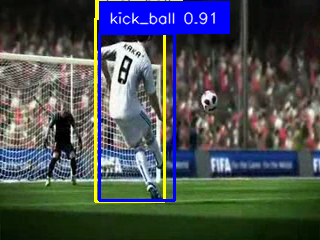} &
\includegraphics[width=\framewidth]{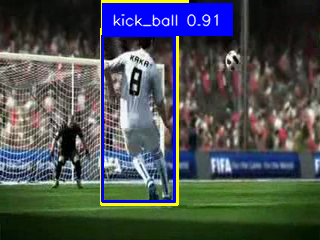} &
\includegraphics[width=\framewidth]{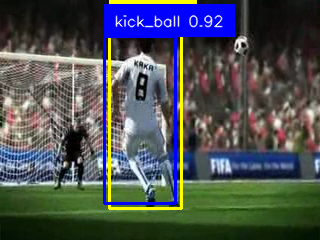}
\\
\parbox[c]{4mm}{\multirow{1}{*}[2.5em]{\rotatebox[origin=c]{90}{pullup}}} &
\includegraphics[width=\framewidth]{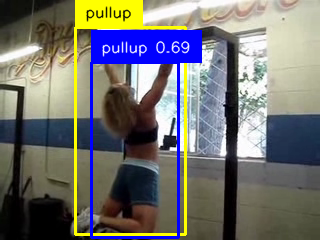} &
\includegraphics[width=\framewidth]{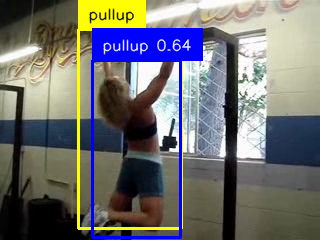} &
\includegraphics[width=\framewidth]{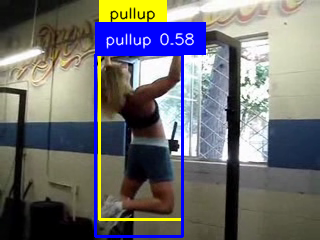} &
\includegraphics[width=\framewidth]{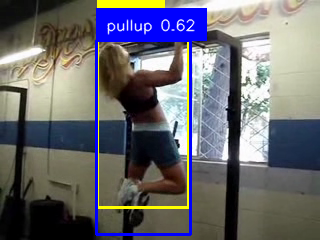} &
\includegraphics[width=\framewidth]{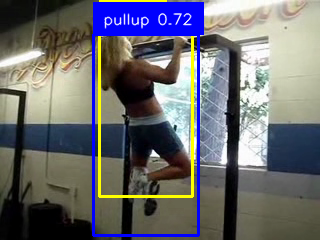} &
\includegraphics[width=\framewidth]{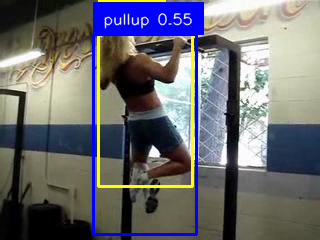}
\\
\end{tabular}
\caption{\textbf{Unseen Action Detection.} We visualize our OpenMixer detections (in \textcolor{blue}{blue}) and ground truth (in \textcolor{yellow}{yellow}) on two representative videos from \textbf{novel} classes. The numbers after class names are confidence scores. More visualizations are in~\cref{supsec:vis}.
}
\label{fig:vis}
\vspace{-6pt}
\end{figure*}

%% file: arXiv_supp.tex
\begin{appendix}

\section*{Appendix}

\section{Prompts for Query-Text Alignment}
\label{supsec:prompts}

To generate text prompts for each action category, we send a request to GPT~\cite{sup-GPT_arXiv2023} by using the template: ``\texttt{For the action type \{CLS\}, what are the visual descriptions? Please respond with a list of 16 short sentences.}" where the placeholder ``\texttt{\{CLS\}}" is replaced by the action class name from the vocabulary. Thus, we obtained multiple caption-like sentence descriptions of the action. Eventually, the text feature for each class is computed by mean pooling of features from the VLM text encoder given the text prompts. In Fig.~\ref{fig:prompt_jhmdb} and Fig.~\ref{fig:prompt_ucf24}, we show a few pieces of the prompt examples on the J-HMDB and UCF101-24 datasets, respectively. We will release all the prompts we used in this work.

\section{Explanation of the Reversed Attention}
\label{appd:reverse_attn}

As discussed in the main paper, the seemly counterintuitive phenomenon of the reversed visual-text attention has been studied in~\cite{sup-RITSM,sup-CLIPSurgery} and we also observed this in our video-based experiments. For CLIP-based models, $\texttt{[CLS]}$ token in ViT is aligned to the text semantics so that its attention weight corresponds to the foreground, while the rest $L$ visual token weights are \underline{complementary} after softmax over $L+1$ tokens before attention pooling. Therefore, due to the attention pooling, high similarity between text feature (or visual $\texttt{[CLS]}$ token feature) and $L$ visual tokens could indicate the background.

\section{Implementation Details}
\label{supsec:impl}

\textbf{Positional Embedding Interpolation.} When using the pre-trained VLM without fine-tuning, an immediate challenge is that the input videos have different spatiotemporal resolutions from the data in VLM pre-training. For example, the CLIP-ViP is pre-trained on input videos with size $12\times 224\times 224$ while videos from J-HMDB can be in any resolution after random augmentations in training. A simple solution is to resize the input video size to match with the pre-trained ones. But for the action detection subtask, person localization is sensitive to the input resolution. To handle this challenge, we instead keep the raw resolution as input, but interpolate the pre-trained spatial and temporal positional embeddings. For example, given the CLIP-ViP B16 VLM and an input video with size $T\times H\times W$, we interpolate the $12$ temporal positional embeddings $\text{PE}_t\in\mathbb{R}^{12\times D}$ to $\hat{\text{PE}}_t\in\mathbb{R}^{T\times D}$, and interpolate the $196\; (=\frac{224}{16}\times\frac{224}{16})$ spatial positional embeddings $\text{PE}_s\in\mathbb{R}^{196\times D}$ to $\hat{\text{PE}}_s\in\mathbb{R}^{L\times D}$ where $L=\frac{H}{16}\times\frac{W}{16}$. This technique is found useful for the action detection problem.

\vspace{1mm}
\noindent\textbf{4D Feature Pyramid.} Following the line of detection literature~\cite{sup-ViTDet_ECCV22,sup-STMixer_CVPR23}, the pre-trained patch token features are transformed into a 4D feature pyramid before the detection head. Let the $\mathbf{H}\in\mathbb{R}^{h\times w\times T\times D}$ be the pre-trained patch token features from the VLM video encoder, where $h\times w$ is the number of patches for each frame, $T$ is the number of video frames, and $D$ is the Transformer dimension. We use deconvolution or convolution to produce hierarchical feature maps $\tilde{\mathbf{H}}^{(l)}$ by spatial strides $s^{(l)} \in \{1/4, 1/2, 1, 2\}$ where the fractional strides are deconvolutional stides and $l$ indexes the pyramid level. Different from~\cite{sup-ViTDet_ECCV22,sup-STMixer_CVPR23} that fully fine-tunes the visual encoder, our VLM visual encoder has to be frozen. Therefore, to allow pre-trained features better utilized by OpenMixer head, we propose to add residual connection at each level of the 4D feature pyramid by spatial interpolation: $\hat{\mathbf{H}}^{(l)} = \phi(\mathbf{H}, s^{(l)}) + \tilde{\mathbf{H}}^{(l)}$. The function $\phi$ is to spatially interpolate the feature map from the size $h\times w$ to the same resolution of $\tilde{\mathbf{H}}^{(l)}$.

\section{Additional Results}
\label{supsec:addres}

\input{tables/vlms}
\input{tables/gptloc}
\input{tables/detector}

\vspace{1mm}
\noindent\textbf{Impact of VLMs.} We note there is a line of literature~\cite{sup-ActionCLIP_arXiv21,sup-rathod2022open_bmvc22,sup-ju2022prompting,sup-nag2022zero,sup-STAN_CVPR23} built on image CLIP for open-vocabulary video understanding. Therefore, it is interesting to see whether image CLIP also works for the OVAD task. In~\cref{tab:vlm}, we compare OpenMixer with its variants using video-based CLIP-ViP~\cite{sup-CLIPViP_ICLR2023} and image-based CLIP~\cite{sup-CLIP_ICML21} under the same ViT-B/16 architecture. The results show that the OpenMixer with CLIP performs way worse than the model with CLIP-ViP, because of the limited capacity of image CLIP in capturing video actions. 

\vspace{1mm}
\noindent\textbf{Can GPT help temporal action localization?} This question is interesting as how textual prompts from language models like GPT could help temporal localization has not been explored in literature. In~\cref{tab:gptloc}, we show that by evaluating the temporal action localization performance, GPT prompts could significantly help.

\vspace{1mm}
\noindent\textbf{Impact of person detectors.} In~\cref{tab:det}, we compare the impact of using external person boxes from off-the-shelf person detectors, \ie, G-DINO~\cite{sup-gdino} and Mask RCNN~\cite{sup-maskrcnn}, in test time on the two best-performed models under the ZSR+ZSL and E2E settings, respectively. It shows that the high-quality boxes from G-DINO could consistently outperform those from Mask RCNN. With the same external test-time boxes, the results of OpenMixer model are consistently better than those of the strongest ZSR+ZSL baseline (Video+GPT). The relatively smaller gains on UCF101-24 than the gains on J-HMDB can be explained by the background bias in UCF videos that restricts VLMs in action recognition.

\input{tables/noise}
\input{tables/gzsl}

\vspace{1mm}
\noindent\textbf{Impact of location prior noise.} In Table~\ref{tab:noise}, we compare ours with 3 variants that use location priors from \textbf{(a)} ground truth (G.T.) boxes which can be regarded as clean without noise and upper-bound (UB) the performance, \textbf{(b)} detected person boxes that may be moderately noisy, and \textbf{(c)} uniform random boxes that are completely noisy and lower-bounds (LB) the performance. The results show our location priors, which are sampled from the text-patch attention map, perform much better than the baselines (b)(c), and are close to the upper-bound performance in (a).

\vspace{1mm}
\noindent\textbf{Generalized zero shot testing.} In our main paper, the base and novel categories are individually given in testing. Thus, in Table~\ref{tab:gtest}, we additionally present the results of the generalized zero-shot testing, in which a complete vocabulary of base and novel categories is given for each testing video. This is more challenging but our OpenMixer still keeps outperformance than the STMixer baseline~\cite{sup-STMixer_CVPR23}. Moreover, according to~\cite{sup-OVRCNN_CVPR21,sup-RegionCLIP_CVPR22}, the rankings of models are stable by the two testing protocols, and only the scales of numbers are different. Therefore, the efficacy of models can still be validated by individual testing in our main paper.

\vspace{1mm}
\noindent\textbf{Results on Different Splits.} We experiment with five random 50\%-50\% seen-unseen class splits on both the J-HMDB and UCF101-24 datasets. Full results of video mAP are summarized in~\cref{tab:splits} and~\ref{tab:splits_ucf}. The split (0) is used in all experiments of the main paper. We also experiment with five random 75\%-25\% seen-unseen class splits on the two datasets, and report results in~\cref{tab:splits75} and~\ref{tab:splits75_ucf}. As some of human actions are much harder to detect than others and they could be included into either base or novel categories, it is normal that the overall performances on different splits vary significantly. Following the existing literature, we will release all splits.
\input{tables/splits}
\input{tables/splits75}

\balance

\section{Visualizations}
\label{supsec:vis}

We present more visualizations on the J-HMDB dataset and UCF101-24 in Fig.~\ref{fig:vis_jhmdb} and~\ref{fig:vis_ucf24}, respectively. They show that our method could detect human actions with precise bounding boxes for both seen and unseen actions. Specifically, in scenarios where multiple persons exist, for the examples of the seen action \texttt{Volleyball Spiking} and the unseen action \texttt{Ice Dancing} on the UCF101-24 dataset, our method could still localize the action-relevant persons on most frames. Referring to single-person action detection, there is still room to improve the performance of multi-person action detection in the future.

\section{Comparison with Concurrent Work~\cite{wu2024open}}
\label{supsec:comp}

The prior work~\cite{wu2024open} defines the same task setting and identifies similar challenges as ours. However, there are several important differences in terms of technical motivations and design. First, for the roadmap, \cite{wu2024open} focuses on large-scale video region-text pre-training followed by downstream fine-tuning, while we emphasize the model adaptation to small downstream datasets in one-time training. Second, for model design,~\cite{wu2024open} is a two-stage method with region proposal generation and action detection refinement, while we adopt DETR-like end-to-end design. As for empirical comparison, currently, this is not feasible because (1) the~\cite{wu2024open} is a concurrent work as ours without releasing any code, data, and models (during the submission period), and (2) it is not an apple-to-apple comparison since the data splits and evaluation metrics of the benchmarks in~\cite{wu2024open} are different from ours as indicated in the paper~\cite{wu2024open}. 

\input{figures/prompts}

\input{figures/demos-suppl/jhmdb/vis}
\input{figures/demos-suppl/ucf24/vis}

\end{appendix}

%% file: tables/vlms.tex
\begin{table}[t]
    \centering
    \setlength{\tabcolsep}{2.5mm}
    \begin{tabular}{l|c|ccc}
         \toprule
         VLMs & Modality  & Mean & Base  & Novel \\
         \midrule
         CLIP~\cite{sup-CLIP_ICML21} & image & 71.60  & 79.46  & 64.44 \\
         CLIP-ViP~\cite{sup-CLIPViP_ICLR2023} &  video & \textbf{86.34} & \textbf{90.75} & \textbf{82.33} \\
         \bottomrule
    \end{tabular}
    \caption{\textbf{Effect of VLMs.} We implement the OpenMixer by CLIP-ViP and CLIP with the same ViT-B/16 transformer.}
    \label{tab:vlm}
\end{table}

%% file: tables/gptloc.tex
\begin{table}[t]
    \centering
    \setlength{\tabcolsep}{4mm}
    \begin{tabular}{l|ccc}
         \toprule
           & Mean(t) & Base(t)  & Novel(t) \\
         \midrule
         w/o. GPT & 83.57   & 90.74   & 77.06 \\
         w. GPT & \textbf{91.62} & \textbf{93.63} & \textbf{89.79} \\
         \bottomrule
    \end{tabular}
    \caption{\textbf{GPT help temporal localization.} We compute mAP by only using temporal IoU on J-HMDB dataset.}
    \label{tab:gptloc}
\end{table}

%% file: tables/detector.tex
\begin{table}[ht]
    \centering
    \footnotesize
    \setlength{\extrarowheight}{0.5mm}
    \setlength{\tabcolsep}{0.8mm}
    \setlength{\abovecaptionskip}{5pt} 
    \begin{tabular}{l|l|ccc|ccc}
         \toprule
         \multirow{2}{*}{models} & \hfil\multirow{2}{*}{person boxes}\hfill & \multicolumn{3}{c|}{J-HMDB} & \multicolumn{3}{c}{UCF101-24} \\
         \cline{3-8}
         & & Mean & Base & Novel   & Mean & Base & Novel \\
         \hline
         \multirow{2}{*}{\shortstack[c]{{\footnotesize ZSR+ZSL}}} 
         & MaskRCNN~\cite{sup-maskrcnn} & 66.73  & 64.61  & 68.66  & 35.01  & 34.59  & 35.43   \\
         & G-DINO~\cite{sup-gdino} & 69.72   & 67.09  & 72.12   & 45.43  & 44.82  & 46.04   \\
         \midrule
         \multirow{2}{*}{\shortstack[c]{{\footnotesize E2E}}} 
         & Mask RCNN~\cite{sup-maskrcnn}  & 83.51  & 87.45  &  79.92 
         & 42.31  & 48.48  & 36.13\\  
         & G-DINO~\cite{sup-gdino}  & 85.06  & 87.76  & 82.60  
         & 46.56  & 47.00  & 46.11  \\  
    \bottomrule
    \end{tabular}
    \caption{\textbf{Impact of person detectors.} For E2E setting, predicted boxes from OpenMixer are replaced with boxes from Mask RCNN~\cite{sup-maskrcnn} or G-DINO~\cite{sup-gdino}, and their classification scores are assigned by maximum IoU with OpenMixer boxes that have scores.}
    \label{tab:det}
\end{table}

%% file: tables/noise.tex
\begin{table}[t]
    \centering
    \setlength{\tabcolsep}{1.5mm}
    \caption{\textbf{Impact of the location priors noise.} We analyze the performance impact from the noise level aspect of the location prior for initializing the box queries of the first S-OMB block.}
    \begin{tabular}{c|l|c|ccc}
         \toprule
         & priors from & noise level & Mean & Base  & Novel \\
         \midrule
         (a) & G.T. (UB) & clean &  91.19   & 93.23   &  89.34   \\
         (b) & detection  &  moderate  &  83.92  & 88.19  &  80.03 \\
         (c) & random (LB) & serious & 54.15   & 56.50    &  52.02   \\
         ours & attention map & slight & 86.34  & 90.75  & 82.33   \\
         \bottomrule
    \end{tabular}
    \label{tab:noise}
\end{table}

%% file: tables/gzsl.tex
\begin{table}[t]
    \centering
    \setlength{\tabcolsep}{1mm}
    \caption{\textbf{Generalized zero-shot testing}. A complete vocabulary of base and novel categories is given in testing.}
    \begin{tabular}{l|ccc|ccc}
         \toprule
          \multirow{2}{*}{Models}\hfill & \multicolumn{3}{c|}{J-HMDB} & \multicolumn{3}{c}{UCF101-24} \\
         \cline{2-7}
         & Mean & Base & Novel   & Mean & Base & Novel \\
         \midrule
         STMixer~\cite{sup-STMixer_CVPR23} & 36.26  & 55.71  & 18.57   & 28.72  & 53.42  & 4.02   \\
         OpenMixer & 74.28  & 77.72  & 71.16   & 40.07  & 54.00  & 26.14   \\
         \bottomrule
    \end{tabular}
    \label{tab:gtest}
\end{table}

%% file: tables/splits.tex
\begin{table}[t]
    \centering
    \setlength{\tabcolsep}{1.5mm}
    \setlength{\extrarowheight}{0.5mm}
    \begin{tabular}{l|ccccc|c}
         \toprule
         Metrics & (0) & (1) & (2) & (3)  & (4) & avg 
         \\
         \hline
         Mean & 86.34  & 86.29  & 85.50  & 86.73  & 83.40 & 85.65 
         \\
         Base & 90.75  & 89.89  & 89.20  & 87.70  & 85.36 & 88.58
         \\
         Novel & 82.33  & 83.02  & 82.13  & 85.85  & 81.61 & 82.99
         \\
         \bottomrule
    \end{tabular}
    \caption{
    \textbf{Results on 50\%-50\% J-HMDB splits.}
    }
    \label{tab:splits}
\end{table}

\begin{table}[t]
    \centering
    \setlength{\tabcolsep}{1.5mm}
    \setlength{\extrarowheight}{0.5mm}
    \begin{tabular}{l|ccccc|c}
         \toprule
         Metrics & (0) & (1) & (2) & (3)  & (4) & avg 
         \\
         \hline
         Mean & 46.42  & 46.28   & 45.45   & 47.32   & 48.30   & 46.75
         \\
         Base & 59.10  & 61.11   & 55.85   & 62.33   & 61.25   & 59.93
         \\
         Novel & 33.73  & 31.45   & 35.05   & 32.31   & 35.34   & 33.58
         \\
         \bottomrule
    \end{tabular}
    \caption{
    \textbf{Results on 50\%-50\% UCF101-24 splits.}
    }
    \label{tab:splits_ucf}
\end{table}

%% file: tables/splits75.tex
\begin{table}[t]
    \centering
    \setlength{\tabcolsep}{1.5mm}
    \setlength{\extrarowheight}{0.5mm}
    \begin{tabular}{l|ccccc|c}
         \toprule
         Metrics & (0) & (1) & (2) & (3)  & (4) & avg 
         \\
         \hline
         Mean & 75.96   & 79.43   & 79.77   &  81.88  &  86.56  & 80.72 
         \\
         Base & 74.73   & 75.21   & 78.34   & 82.14   &  85.46  & 79.17 
         \\
         Novel & 79.03   & 89.98   & 83.34   & 81.23   & 89.30   & 84.57 
         \\
         \bottomrule
    \end{tabular}
    \caption{
    \textbf{Results on 75\%-25\% J-HMDB splits.}
    }
    \label{tab:splits75}
\end{table}

\begin{table}[t]
    \centering
    \setlength{\tabcolsep}{1.5mm}
    \setlength{\extrarowheight}{0.5mm}
    \begin{tabular}{l|ccccc|c}
         \toprule
         Metrics & (0) & (1) & (2) & (3)  & (4) & avg 
         \\
         \hline
         Mean & 55.78   & 55.83   & 57.04   & 57.19   & 61.85   & 57.54 
         \\
         Base & 64.85   & 61.83   & 60.16   & 58.74   & 61.82   & 61.48 
         \\
         Novel & 28.55   & 37.80   & 47.69   & 52.55   & 61.96   & 45.71 
         \\
         \bottomrule
    \end{tabular}
    \caption{
    \textbf{Results on 75\%-25\% UCF101-24 splits.}
    }
    \label{tab:splits75_ucf}
\end{table}

%% file: figures/prompts.tex
\begin{figure*}[t]
\begin{lstlisting}[breakatwhitespace=false]
{
"brush_hair": "Brush Hair: A person is brushing their hair using hand movements with a hairbrush or their fingers.",
"catch": "Catch: Someone is catching an object, such as a ball or a frisbee, with their hands.",
"clap": "Clap: A person is bringing their hands together to create a clapping sound.",
"climb_stairs": "Climb Stairs: Someone is ascending or descending a set of stairs, using alternating leg movements.",
"golf": "Golf: A person is swinging a golf club to hit a golf ball.",
"jump": "Jump: Someone is propelling themselves off the ground using both feet simultaneously.",
"kick_ball": "Kick Ball: A person is striking a ball with their foot.",
"pick": "Pick: Someone is picking up an object from the ground, usually using their hands.",
"pour": "Pour: A person is pouring liquid from one container to another.",
"pullup": "Pull Up: Someone is lifting their body upwards using their arms, typically performed on a horizontal bar.",
"push": "Push: A person is exerting force on an object away from their body, using their hands or body.",
"run": "Run: Someone is moving quickly on their feet, usually in a straight line.",
"shoot_ball": "Shoot Ball: A person is shooting a ball towards a target or a goal using their hands or feet.",
"shoot_bow": "Shoot Bow: Someone is using a bow to shoot an arrow.",
"shoot_gun": "Shoot Gun: A person is firing a gun, typically aimed at a target.",
"sit": "Sit: Someone is in a seated position with their weight supported by a surface, such as a chair.",
"stand": "Stand: A person is upright on their feet, with their body fully supported by their legs.",
"swing_baseball": "Swing Baseball Bat: Someone is swinging a baseball bat to hit a ball.",
"throw": "Throw: A person is propelling an object through the air using their hand or arm.",
"walk": "Walk: Someone is moving on their feet with a regular, steady pace, but slower than running.",
"wave": "Wave: A person is moving their hand or arm back and forth in a greeting or farewell gesture, usually with an open palm."
}
\end{lstlisting}
\caption{Generated prompts for J-HMDB action categories. For each category, we generate one prompt sentence.}
\label{fig:prompt_jhmdb}
\end{figure*}

\begin{figure*}[t]
\begin{lstlisting}[breakatwhitespace=false]
{"Basketball": [
    "Basketball: A player dribbles the ball swiftly down the court amidst cheers from the crowd.",
    "Basketball: An athlete performs a high jump and slam dunks the ball into the net with confidence.",
    "Basketball: Teammates pass the ball around the court, strategizing their next move.",
    "Basketball: A player precision shoots the ball from the three-point line and scores.",
    "Basketball: A tense one-on-one standoff as a player attempts to steal the ball.",
    "Basketball: Players execute deft maneuvers around opponents on the court.",
    "Basketball: A player displays impressive footwork while maintaining control of the ball.",
    "Basketball: Following a whistle blow, a player steps up to take a free throw.",
    "Basketball: The coach calls a timeout to relay new strategies to the team.",
    "Basketball: A swift breakaway leads to a stunning layup and two points on the board.",
    "Basketball: Thorny defense put up by players trying to prevent the opposing team from scoring.",
    "Basketball: The player manages to steal the ball, intercepting a pass and turning the game around.",
    "Basketball: In the sound of the last buzzer, players celebrate a well-earned victory.",
    "Basketball: Spectators erupt in cheers as the ball swishes through the net.",
    "Basketball: A captivating display of agility and teamwork witnessed on the court.",
    "Basketball: A player makes a long, arching shot from the half-court line, electrifying the crowd."
],
.
.
.
"TrampolineJumping": [
    "Trampoline Jumping: A joyful child is leaping high on a trampoline in their backyard.",
    "Trampoline Jumping: A gymnast is skillfully performing somersaults on a trampoline.",
    "Trampoline Jumping: A group of friends are competing in tricks while bouncing on a trampoline.",
    "Trampoline Jumping: A professional athlete is executing a perfect backflip on a trampoline. ",
    "Trampoline Jumping: Enthralled family members are enjoying a trampoline jump session on a sunny day.",
    "Trampoline Jumping: Excited children are bouncing and laughing on a trampoline at a birthday party.",
    "Trampoline Jumping: A fitness enthusiast is getting an intense workout by jumping on a trampoline.",
    "Trampoline Jumping: An acrobat rehearses complicated maneuvers on a large trampoline. ",
    "Trampoline Jumping: A fearless teenager is executing high jumps on a trampoline in a skate park.",
    "Trampoline Jumping: An adventurous person is defying gravity with bounces on a massive trampoline.",
    "Trampoline Jumping: A young girl confidently performs flips and twists on a trampoline. ",
    "Trampoline Jumping: A trampoline athlete practices precise landings in a professional gym.",
    "Trampoline Jumping: An aspiring gymnast is perfecting their routine on a trampoline.",
    "Trampoline Jumping: A boy exhilaratingly jumps towards the sky on a trampoline, his laughter filling the air.",
    "Trampoline Jumping: A daring young woman is doing mid-air splits on a trampoline in an indoor park.",
    "Trampoline Jumping: A man is reaching extreme heights, all while being propelled off a trampoline."
]}
\end{lstlisting}
\caption{Generated prompts for UCF101-24 action categories. For each category, we generate 16 prompt sentences.}
\label{fig:prompt_ucf24}
\end{figure*}

%% file: figures/demos-suppl/jhmdb/vis.tex
\begin{figure*}[t]
\scriptsize
\centering
\renewcommand{\tabcolsep}{0.7pt} %
\begin{tabular}{ccccccc}
& $t/T=1/6$ & $t/T=2/6$ & $t/T=3/6$ & $t/T=4/6$ & $t/T=5/6$ & $t/T=1$
\\
\parbox[c]{4mm}{\multirow{1}{*}[4.5em]{\rotatebox[origin=c]{90}{\textbf{brush hair}}}} &
\includegraphics[width=\framewidth]{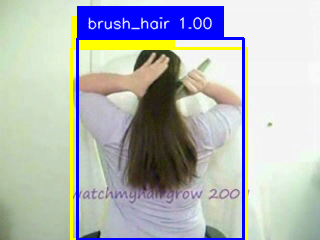} &
\includegraphics[width=\framewidth]{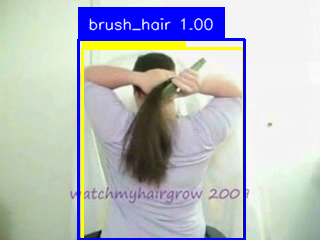} &
\includegraphics[width=\framewidth]{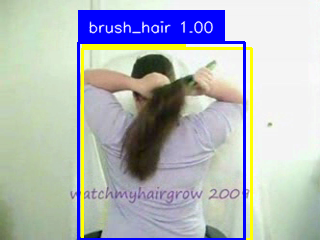} &
\includegraphics[width=\framewidth]{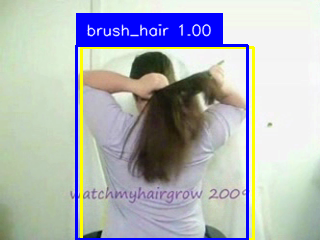} &
\includegraphics[width=\framewidth]{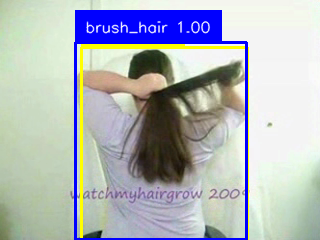} &
\includegraphics[width=\framewidth]{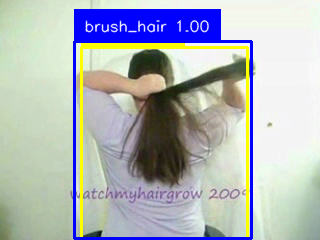}
\\
\parbox[c]{4mm}{\multirow{1}{*}[3.0em]{\rotatebox[origin=c]{90}{\textbf{catch}}}} &
\includegraphics[width=\framewidth]{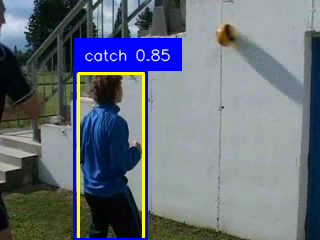} &
\includegraphics[width=\framewidth]{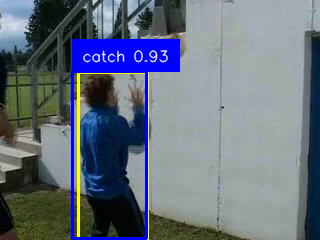} &
\includegraphics[width=\framewidth]{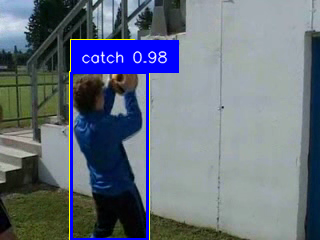} &
\includegraphics[width=\framewidth]{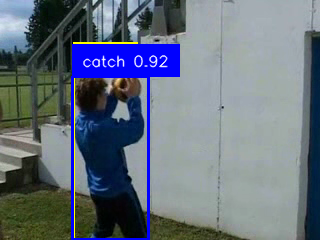} &
\includegraphics[width=\framewidth]{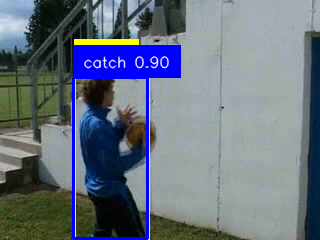} &
\includegraphics[width=\framewidth]{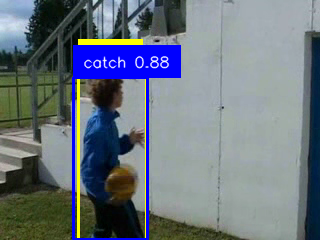}
\\
\parbox[c]{4mm}{\multirow{1}{*}[3.0em]{\rotatebox[origin=c]{90}{\textbf{pick}}}} &
\includegraphics[width=\framewidth]{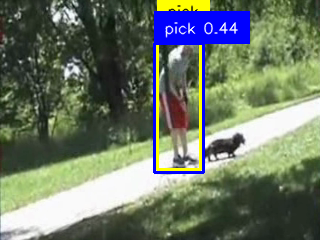} &
\includegraphics[width=\framewidth]{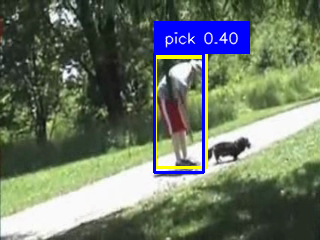} &
\includegraphics[width=\framewidth]{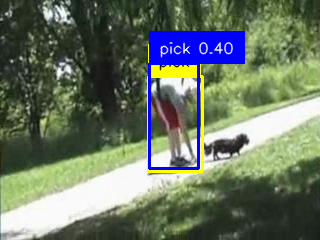} &
\includegraphics[width=\framewidth]{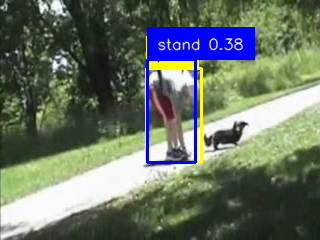} &
\includegraphics[width=\framewidth]{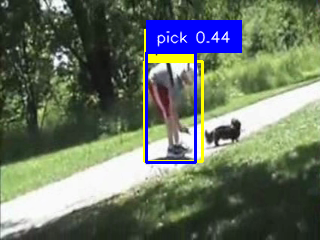} &
\includegraphics[width=\framewidth]{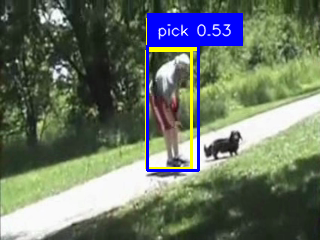}
\\
\parbox[c]{4mm}{\multirow{1}{*}[3.0em]{\rotatebox[origin=c]{90}{\textbf{pour}}}} &
\includegraphics[width=\framewidth]{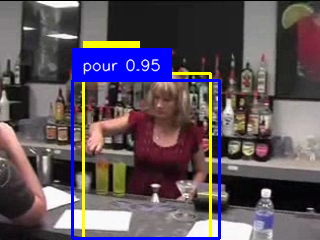} &
\includegraphics[width=\framewidth]{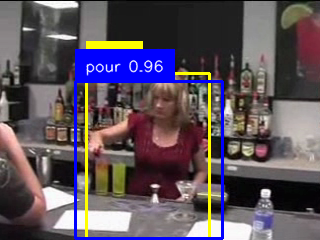} &
\includegraphics[width=\framewidth]{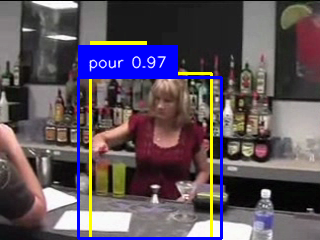} &
\includegraphics[width=\framewidth]{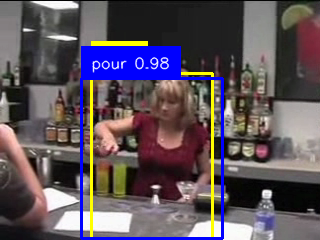} &
\includegraphics[width=\framewidth]{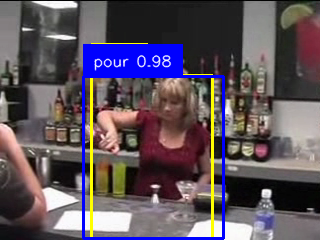} &
\includegraphics[width=\framewidth]{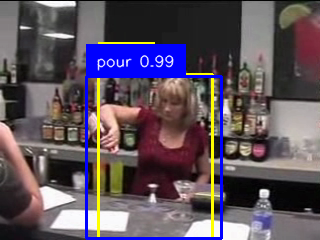}
\\
\parbox[c]{4mm}{\multirow{1}{*}[3.0em]{\rotatebox[origin=c]{90}{\textbf{push}}}} &
\includegraphics[width=\framewidth]{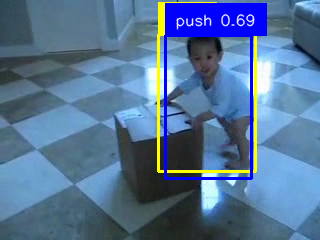} &
\includegraphics[width=\framewidth]{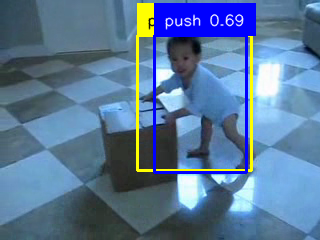} &
\includegraphics[width=\framewidth]{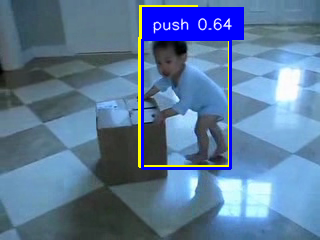} &
\includegraphics[width=\framewidth]{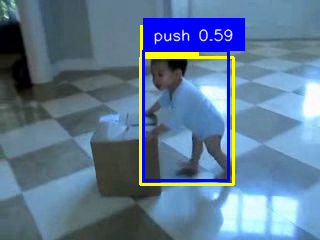} &
\includegraphics[width=\framewidth]{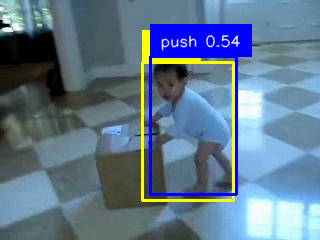} &
\includegraphics[width=\framewidth]{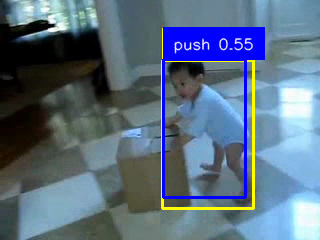}
\\
\parbox[c]{4mm}{\multirow{1}{*}[2.5em]{\rotatebox[origin=c]{90}{\textcolor{red}{\textbf{golf}}}}} &
\includegraphics[width=\framewidth]{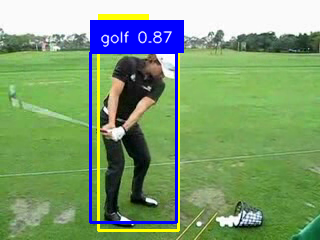} &
\includegraphics[width=\framewidth]{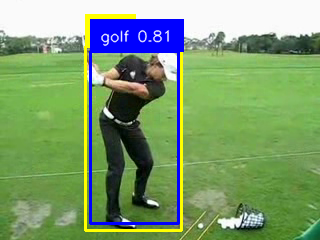} &
\includegraphics[width=\framewidth]{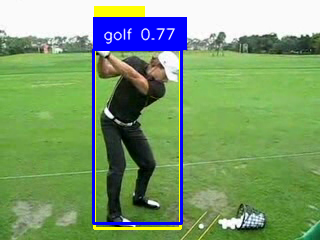} &
\includegraphics[width=\framewidth]{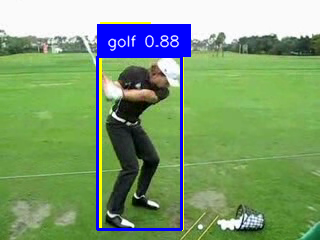} &
\includegraphics[width=\framewidth]{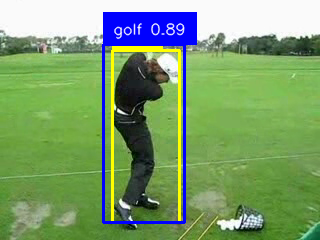} &
\includegraphics[width=\framewidth]{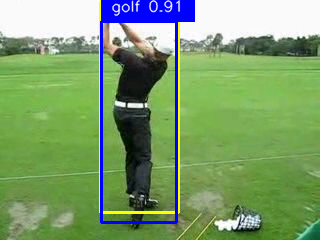}
\\
\parbox[c]{4mm}{\multirow{1}{*}[4.5em]{\rotatebox[origin=c]{90}{\textcolor{red}{\textbf{shoot bow}}}}} &
\includegraphics[width=\framewidth]{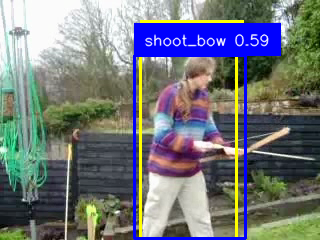} &
\includegraphics[width=\framewidth]{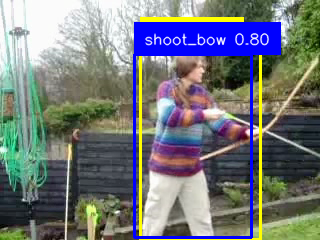} &
\includegraphics[width=\framewidth]{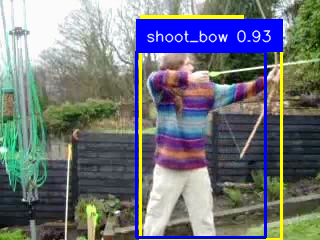} &
\includegraphics[width=\framewidth]{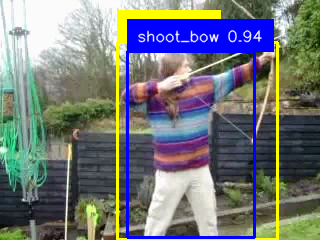} &
\includegraphics[width=\framewidth]{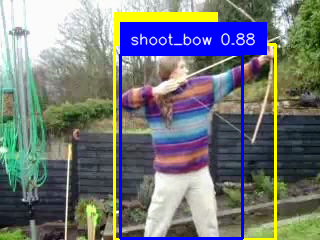} &
\includegraphics[width=\framewidth]{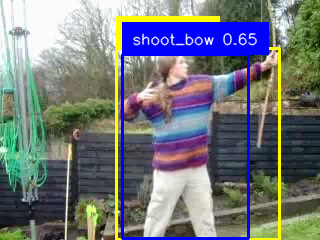}
\\
\parbox[c]{4mm}{\multirow{1}{*}[4.0em]{\rotatebox[origin=c]{90}{\textcolor{red}{\textbf{shoot gun}}}}} &
\includegraphics[width=\framewidth]{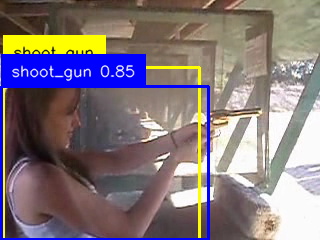} &
\includegraphics[width=\framewidth]{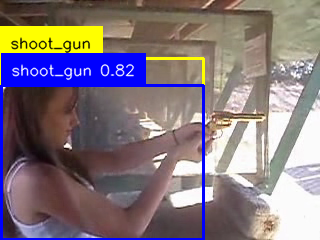} &
\includegraphics[width=\framewidth]{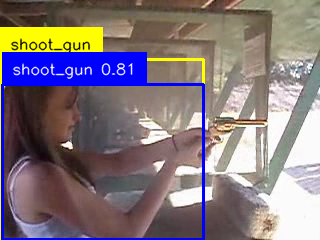} &
\includegraphics[width=\framewidth]{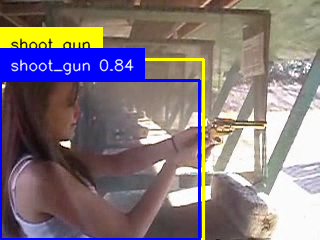} &
\includegraphics[width=\framewidth]{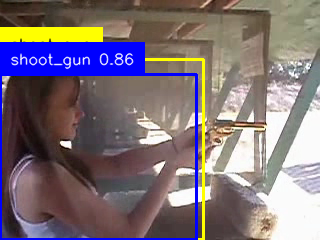} &
\includegraphics[width=\framewidth]{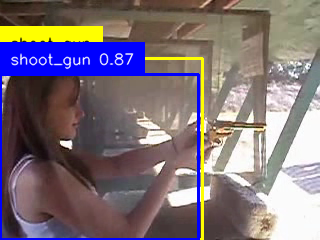}
\\
\parbox[c]{4mm}{\multirow{1}{*}[2.5em]{\rotatebox[origin=c]{90}{\textcolor{red}{\textbf{sit}}}}} &
\includegraphics[width=\framewidth]{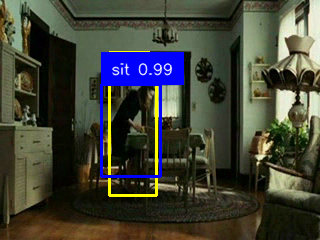} &
\includegraphics[width=\framewidth]{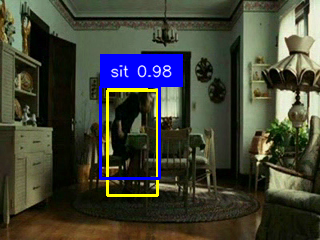} &
\includegraphics[width=\framewidth]{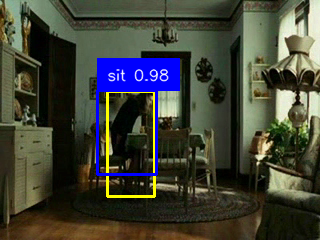} &
\includegraphics[width=\framewidth]{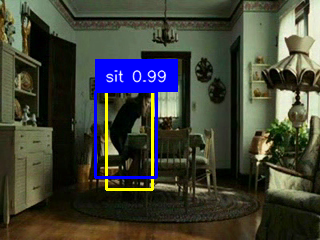} &
\includegraphics[width=\framewidth]{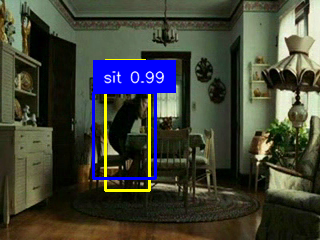} &
\includegraphics[width=\framewidth]{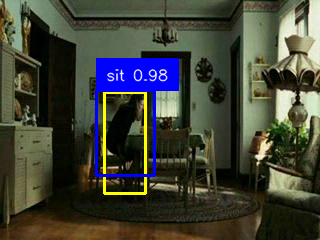}
\\
\parbox[c]{4mm}{\multirow{1}{*}[3.5em]{\rotatebox[origin=c]{90}{\textcolor{red}{\textbf{baseball}}}}} &
\includegraphics[width=\framewidth]{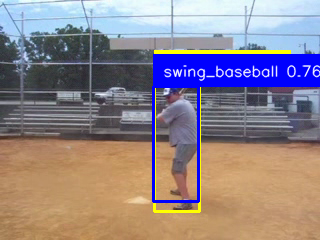} &
\includegraphics[width=\framewidth]{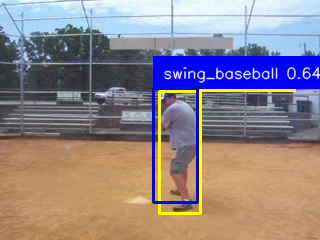} &
\includegraphics[width=\framewidth]{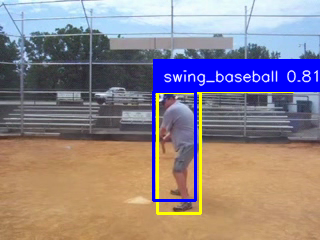} &
\includegraphics[width=\framewidth]{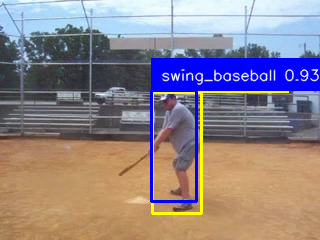} &
\includegraphics[width=\framewidth]{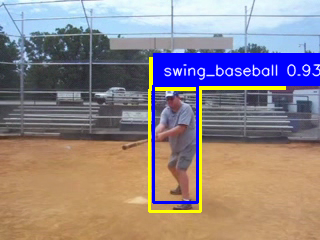} &
\includegraphics[width=\framewidth]{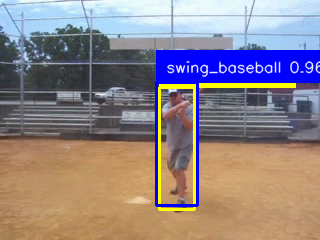}
\\
\end{tabular}
\captionsetup{font=small,aboveskip=3pt}
\caption{\textbf{Visualization on J-HMDB dataset.} We visualize our OpenMixer detections (in \textcolor{blue}{blue boxes}) and ground truth (in \textcolor{yellow}{yellow boxes}) on five base classes (in \textcolor{black}{\textbf{black font}}) and five novel classes (in \textcolor{red}{\textbf{red font}}). Class names are shortened for brevity. The numbers after class names are confidence scores.
}
\label{fig:vis_jhmdb}
\vspace{-6pt}
\end{figure*}

%% file: figures/demos-suppl/ucf24/vis.tex
\begin{figure*}[t]
\tiny
\centering
\renewcommand{\tabcolsep}{0.7pt} %
\begin{tabular}{ccccccc}
& $t/T=1/6$ & $t/T=2/6$ & $t/T=3/6$ & $t/T=4/6$ & $t/T=5/6$ & $t/T=1$
\\
\parbox[c]{4mm}{\multirow{1}{*}[4.0em]{\rotatebox[origin=c]{90}{\textbf{Biking}}}} &
\includegraphics[width=\framewidth]{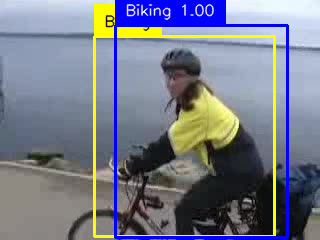} &
\includegraphics[width=\framewidth]{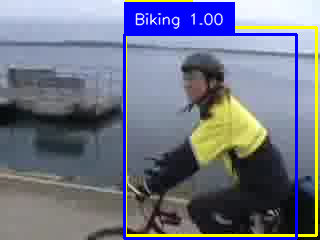} &
\includegraphics[width=\framewidth]{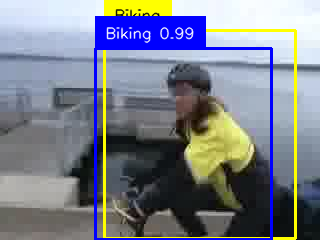} &
\includegraphics[width=\framewidth]{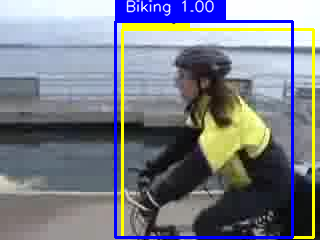} &
\includegraphics[width=\framewidth]{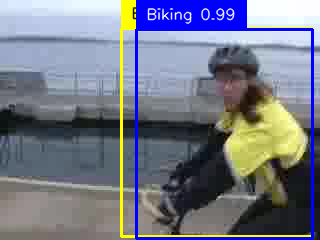} &
\includegraphics[width=\framewidth]{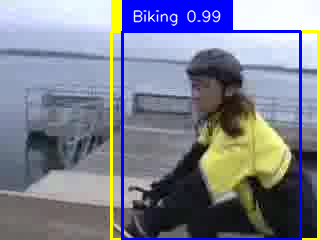}
\\
\parbox[c]{4mm}{\multirow{1}{*}[5.5em]{\rotatebox[origin=c]{90}{\textbf{Floor Gym.}}}} &
\includegraphics[width=\framewidth]{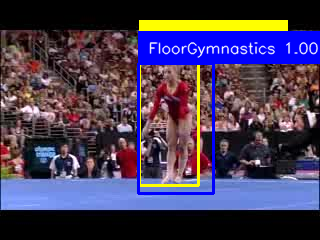} &
\includegraphics[width=\framewidth]{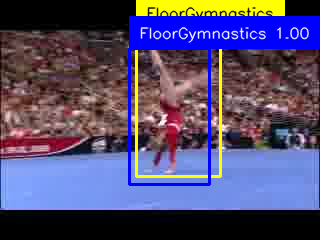} &
\includegraphics[width=\framewidth]{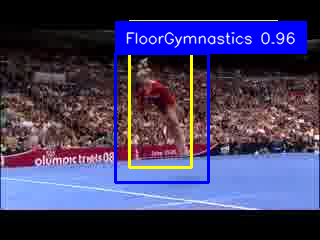} &
\includegraphics[width=\framewidth]{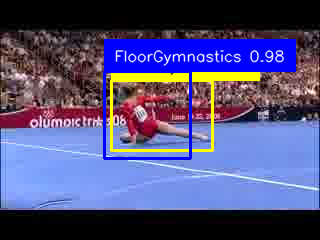} &
\includegraphics[width=\framewidth]{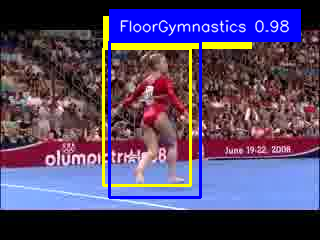} &
\includegraphics[width=\framewidth]{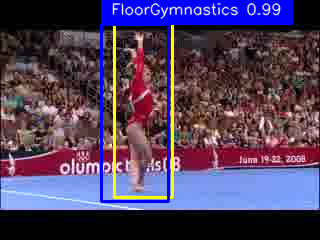}
\\
\parbox[c]{4mm}{\multirow{1}{*}[5.0em]{\rotatebox[origin=c]{90}{\textbf{Horse Riding}}}} &
\includegraphics[width=\framewidth]{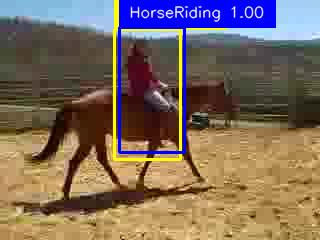} &
\includegraphics[width=\framewidth]{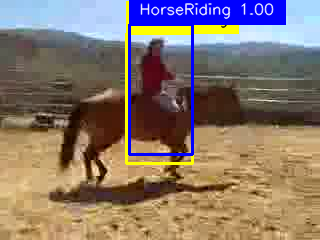} &
\includegraphics[width=\framewidth]{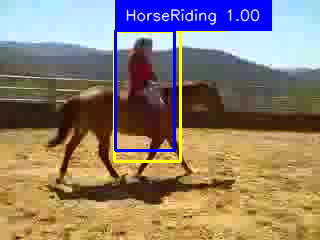} &
\includegraphics[width=\framewidth]{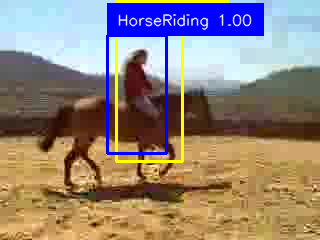} &
\includegraphics[width=\framewidth]{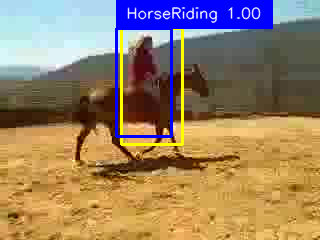} &
\includegraphics[width=\framewidth]{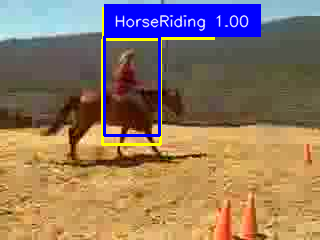}
\\
\parbox[c]{4mm}{\multirow{1}{*}[4.0em]{\rotatebox[origin=c]{90}{\textbf{Surfing}}}} &
\includegraphics[width=\framewidth]{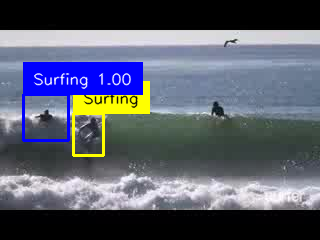} &
\includegraphics[width=\framewidth]{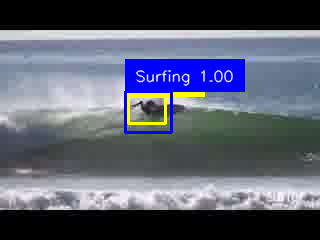} &
\includegraphics[width=\framewidth]{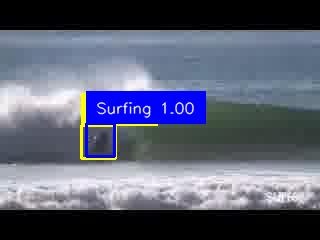} &
\includegraphics[width=\framewidth]{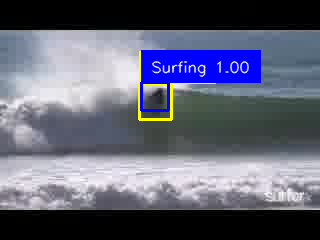} &
\includegraphics[width=\framewidth]{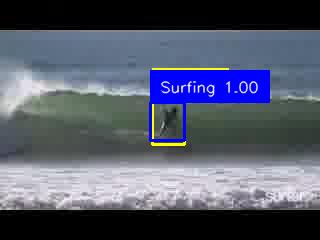} &
\includegraphics[width=\framewidth]{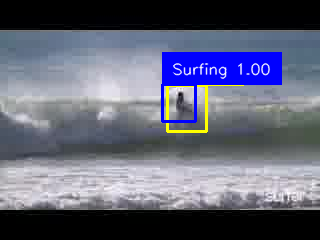}
\\
\parbox[c]{4mm}{\multirow{1}{*}[4.5em]{\rotatebox[origin=c]{90}{\textbf{Volleyball}}}} &
\includegraphics[width=\framewidth]{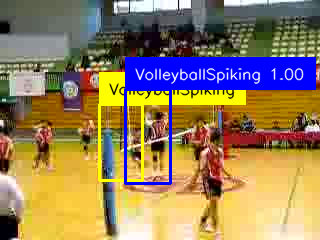} &
\includegraphics[width=\framewidth]{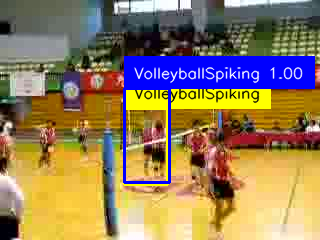} &
\includegraphics[width=\framewidth]{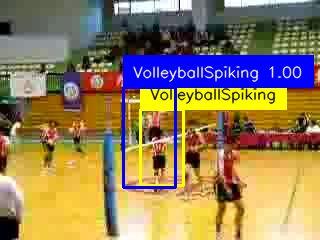} &
\includegraphics[width=\framewidth]{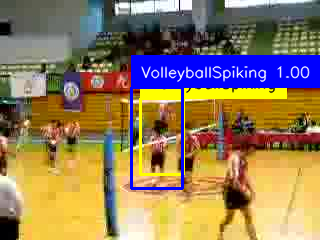} &
\includegraphics[width=\framewidth]{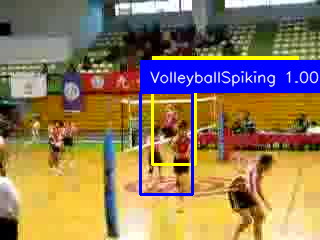} &
\includegraphics[width=\framewidth]{figures/demos-suppl/ucf24/base/VolleyballSpiking/57.png}
\\
\parbox[c]{4mm}{\multirow{1}{*}[4.0em]{\rotatebox[origin=c]{90}{\textcolor{red}{\textbf{Basketball}}}}} &
\includegraphics[width=\framewidth]{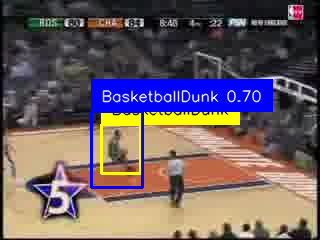} &
\includegraphics[width=\framewidth]{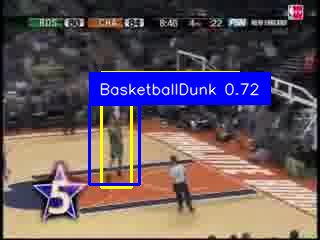} &
\includegraphics[width=\framewidth]{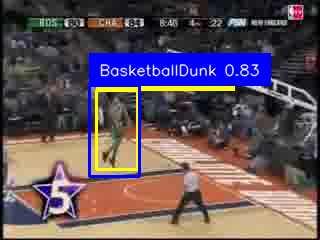} &
\includegraphics[width=\framewidth]{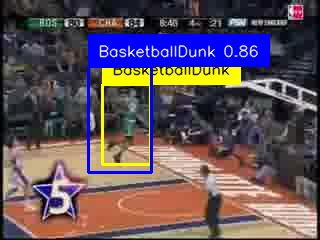} &
\includegraphics[width=\framewidth]{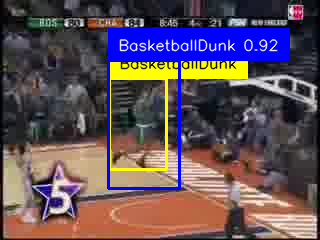} &
\includegraphics[width=\framewidth]{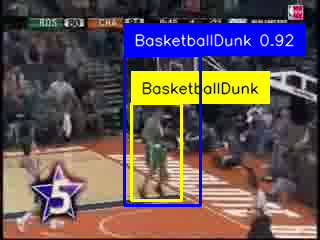}
\\
\parbox[c]{4mm}{\multirow{1}{*}[4.5em]{\rotatebox[origin=c]{90}{\textcolor{red}{\textbf{Ice Dance}}}}} &
\includegraphics[width=\framewidth]{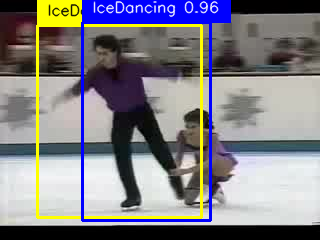} &
\includegraphics[width=\framewidth]{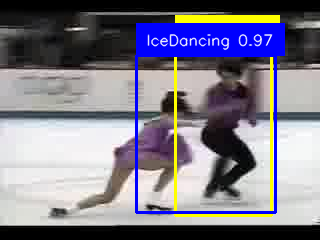} &
\includegraphics[width=\framewidth]{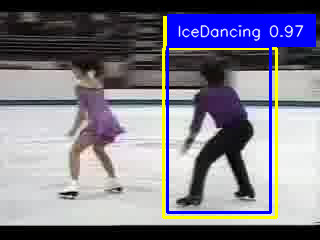} &
\includegraphics[width=\framewidth]{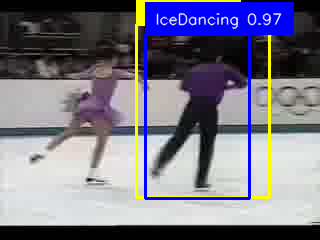} &
\includegraphics[width=\framewidth]{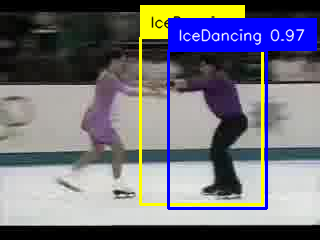} &
\includegraphics[width=\framewidth]{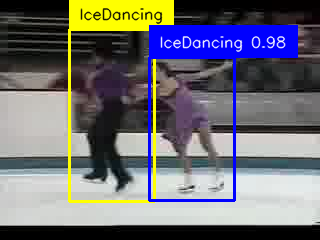}
\\
\parbox[c]{4mm}{\multirow{1}{*}[5.0em]{\rotatebox[origin=c]{90}{\textcolor{red}{\textbf{Long Jump}}}}} &
\includegraphics[width=\framewidth]{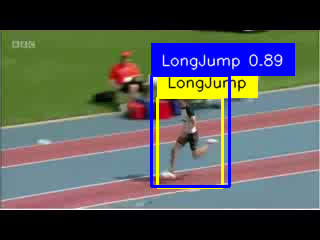} &
\includegraphics[width=\framewidth]{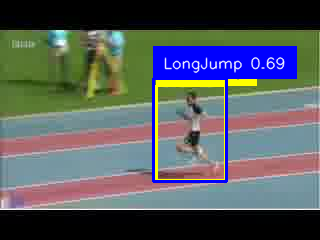} &
\includegraphics[width=\framewidth]{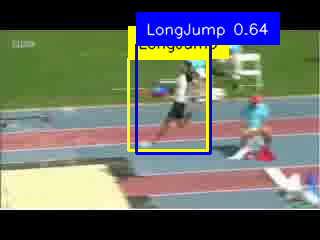} &
\includegraphics[width=\framewidth]{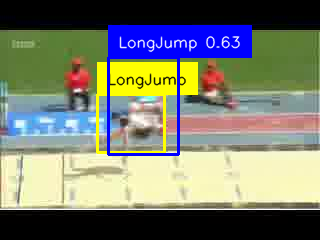} &
\includegraphics[width=\framewidth]{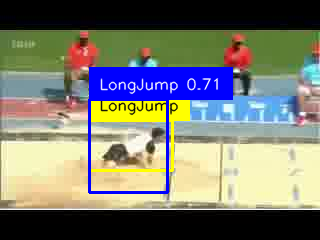} &
\includegraphics[width=\framewidth]{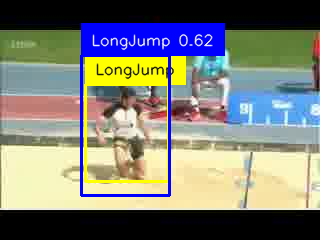}
\\
\parbox[c]{4mm}{\multirow{1}{*}[4.0em]{\rotatebox[origin=c]{90}{\textcolor{red}{\textbf{Skijet}}}}} &
\includegraphics[width=\framewidth]{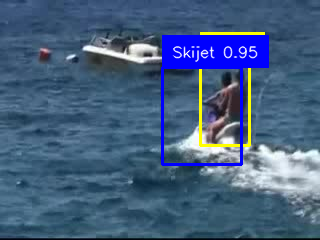} &
\includegraphics[width=\framewidth]{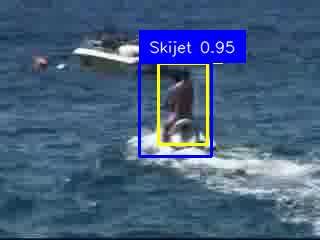} &
\includegraphics[width=\framewidth]{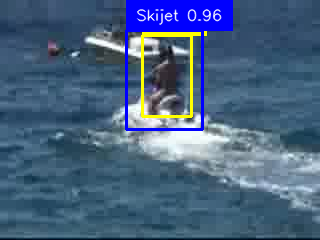} &
\includegraphics[width=\framewidth]{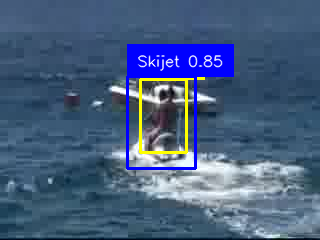} &
\includegraphics[width=\framewidth]{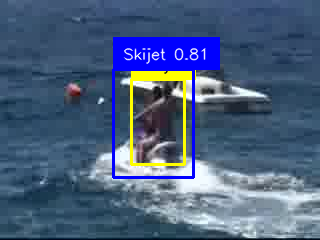} &
\includegraphics[width=\framewidth]{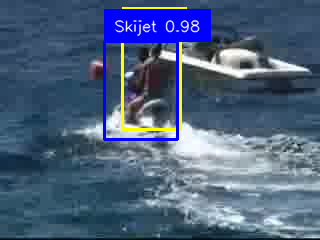}
\\
\parbox[c]{4mm}{\multirow{1}{*}[5.5em]{\rotatebox[origin=c]{90}{\textcolor{red}{\textbf{Walking Dog}}}}} &
\includegraphics[width=\framewidth]{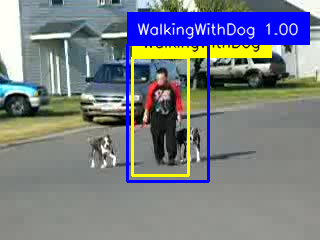} &
\includegraphics[width=\framewidth]{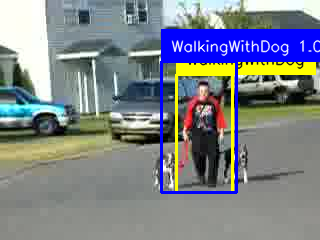} &
\includegraphics[width=\framewidth]{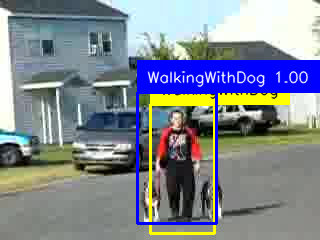} &
\includegraphics[width=\framewidth]{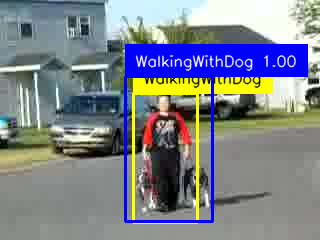} &
\includegraphics[width=\framewidth]{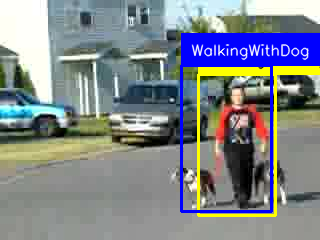} &
\includegraphics[width=\framewidth]{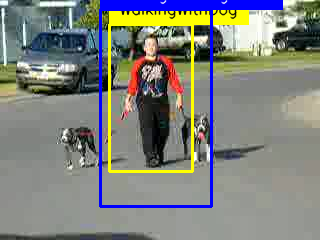}
\\
\end{tabular}
\captionsetup{font=small,aboveskip=3pt}
\caption{\textbf{Visualization on UCF101-24 dataset.} We visualize our OpenMixer detections (in \textcolor{blue}{blue boxes}) and ground truth (in \textcolor{yellow}{yellow boxes}) on five base classes (in \textcolor{black}{\textbf{black font}}) and five novel classes (in \textcolor{red}{\textbf{red font}}). Class names are shortened for brevity. The numbers after class names are confidence scores.
}
\label{fig:vis_ucf24}
\vspace{-6pt}
\end{figure*}